\pdfoutput=1
\documentclass[11pt]{article}

\usepackage[final]{acl}

\usepackage{times}
\usepackage{latexsym}
\usepackage{amsmath}
\usepackage[T1]{fontenc}
\usepackage{booktabs}

\usepackage[utf8]{inputenc}

\usepackage{microtype}

\usepackage{inconsolata}

\usepackage{graphicx}

\usepackage{adjustbox}

\usepackage[table]{xcolor}
\definecolor{light-gray}{gray}{0.94}

\usepackage{amssymb} % for \checkmark
\newcommand{\cmark}{\ensuremath{\checkmark}}
\usepackage{pifont}
\title{A Robust Evaluation of Probe Robustness: \\
Lessons for Reliable OOD Uncertainty Quantification}

\author{Joe Stacey\textsuperscript{1}, 
Hadas Orgad\textsuperscript{2}, 
Kentaro Inui\textsuperscript{3,4,5}, \\
\textbf{Benjamin Heinzerling\textsuperscript{5,4}}, 
\textbf{Nafise Sadat Moosavi\textsuperscript{1}} \\
\textsuperscript{1}University of Sheffield, \textsuperscript{2}Kempner Institute at Harvard University, \\\textsuperscript{3}MBZUAI, \textsuperscript{4}Tohoku University, \textsuperscript{5}RIKEN, \\
\texttt{\{j.stacey, n.s.moosavi\}@sheffield.ac.uk, hadasorgad@fas.harvard.edu}, \\ 
\texttt{kentaro.inui@mbzuai.ac.ae, benjamin.heinzerling@riken.jp}
}

\begin{document}
\maketitle

\begin{abstract}
Recent work has shown that the hidden states of large language models contain signals useful for uncertainty estimation, motivating a growing interest in efficient probe-based approaches. Yet it remains unclear how robust existing methods are, with prior work reporting conflicting conclusions under substantially different evaluation settings. We address this by introducing ProbeDrift, a systematic evaluation framework for supervised uncertainty probes covering a wide range of OOD settings across models, tasks, and distributional shifts. 
Using ProbeDrift, we train over 2,000 probes to disentangle the effect of key design choices, showing poor robustness of current methods beyond near-OOD settings. We find that robustness is driven by design decisions that have a largely invisible effect in-distribution, including the choice of feature type, aggregation strategy, and training signal.
We argue that robust uncertainty estimation requires robust evaluation. To support this, we release ProbeDrift as a lightweight Python library that contains the train and test splits underpinning our extensive evaluation.\footnote{\url{https://github.com/joestacey/ProbeDrift}} We also show how insights from our evaluation can directly lead to more robust methods through a simple Hybrid Back-Off (HBO) strategy.
\end{abstract}

\section{Introduction}

\begin{figure}[t]
  \centering
  \includegraphics[width=\columnwidth]{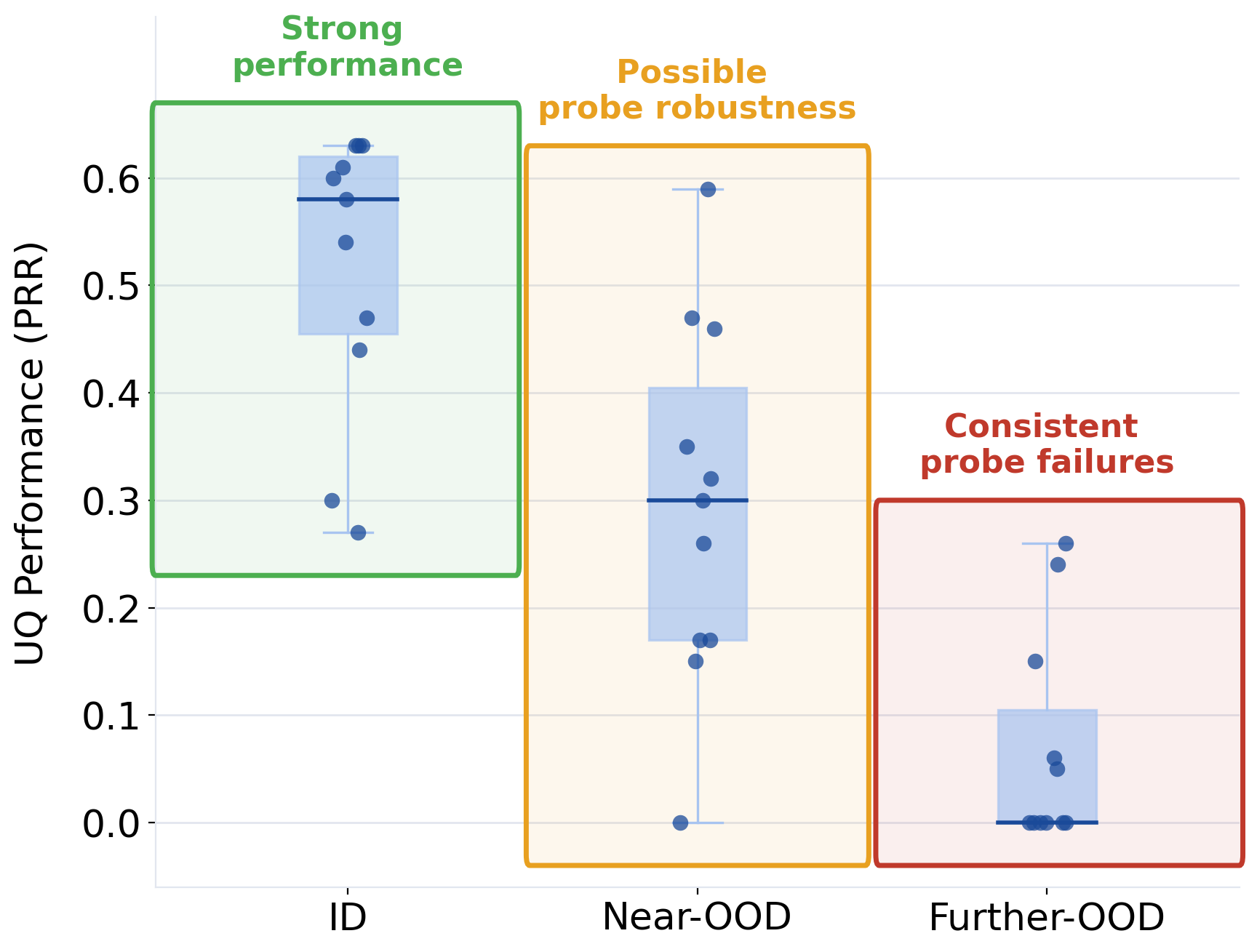}
  \caption{Uncertainty Quantification methods quickly lose robustness outside of their training distribution, and these failures are often not visible in existing OOD evaluation. We show that robustness is possible in a near-OOD setting with the right probe configuration. Results are from Table \ref{tab:short_long_generation_results_two_models} for Llama-3.1-8B, with negative PRR scores floored at 0 for visualisation.}
  \label{fig:summary_diagram}
\end{figure}
Supervised Uncertainty Quantification (UQ) methods train probes to predict the reliability of model outputs from internal features such as hidden states or attention \cite{azaria-mitchell-2023-internal, DBLP:conf/iclr/OrgadTGRSKB25, he-etal-2024-llm, obeso2025realtimedetectionhallucinatedentities}. These UQ methods achieve strong in-distribution performance, providing evidence that model internals contain signals related to correctness and uncertainty. However, there is mixed evidence about the robustness of these probes, with some work reporting strong out-of-distribution (OOD) performance \cite{shelmanov-etal-2025-head, obeso2025realtimedetectionhallucinatedentities, vazhentsev-etal-2025-token} while other work finds poor robustness \cite{Levinstein_2024, DBLP:conf/iclr/OrgadTGRSKB25}. 
Since probing methods differ in their features, architecture, and aggregation strategies, existing OOD findings often do not generalise beyond the specific methods being tested. Different experimental setups, baselines, and definitions of distributional shifts make comparisons between prior works even more challenging. It is therefore unclear whether strong robustness claims reflect genuine improvements or differences in evaluation.

In this work we introduce ProbeDrift, a systematic evaluation framework for measuring the robustness of supervised UQ probes across progressively stronger distributional shifts. ProbeDrift enables consistent comparisons across probe designs, tasks, models, and OOD settings, addressing the fragmented evaluation used in prior work. Using this framework, we train and evaluate over 2,000 probes while varying key design dimensions including representation layer, feature type, training signal, and token aggregation strategy.

Our evaluation reveals that probes can generalise under limited distributional shifts, but this generalisation quickly degrades for larger shifts. 
We find that probes trained on middle-layer hidden states generalise better than those from the final layer, and aggregating information across response tokens is better than relying on single-token features. We also find that the choice of correctness function used during training can substantially affect robustness, particularly for long-form generations. These design choices often have a minimal impact in-distribution, and remain hidden without systematic OOD evaluation.

Ultimately, we argue that robust uncertainty quantification requires robust evaluation. Insufficient evaluation can mask generalisation failures and obscure probe design choices which become critical under larger distributional shifts. In support of this, we provide ProbeDrift as a lightweight Python library to make our extensive evaluation easy to reproduce. We further show how insights from comprehensive evaluation can directly guide the design of more robust methods through a simple Hybrid Back-Off (HBO) strategy which substantially improves robustness.

To summarise our contributions:
\begin{enumerate}
\itemsep-0.2em
\item We introduce ProbeDrift, a systematic evaluation framework for assessing the robustness of supervised UQ probes across progressively stronger distributional shifts. Using ProbeDrift, we comprehensively evaluate existing supervised UQ methods, releasing this as a lightweight Python library to support future work.
\item We disentangle key design factors, including the choice of architecture, feature, correctness function and token aggregation strategy, finding choices with little effect in-distribution can substantially impact robustness.
\item We show that robustness conclusions for supervised UQ probes depend strongly on the evaluation setting, highlighting the need for systematic robustness evaluation across progressively stronger distributional shifts.
\item We introduce Hybrid Back-Off (HBO) as an illustrative example to demonstrate how insights from ProbeDrift can directly guide the development of more robust methods.\footnote{\url{https://github.com/joestacey/RobustUQProbes}}

\end{enumerate}

\section{Background}

There is mixed evidence about the robustness of uncertainty quantification probes. Several recent methods report strong generalisation across tasks and datasets \cite{shelmanov-etal-2025-head, obeso2025realtimedetectionhallucinatedentities, vazhentsev-etal-2025-token}, while other work observes OOD failures \cite{ch-wang-etal-2024-androids, Levinstein_2024, DBLP:conf/iclr/OrgadTGRSKB25}. For instance, \citet{Levinstein_2024} find that probes do not generalise to texts containing negations, while \citet{DBLP:conf/iclr/OrgadTGRSKB25} find that probes rarely outperform unsupervised baselines when testing performance OOD. \citet{ch-wang-etal-2024-androids} report similar findings, showing that an ensemble of probes with linear classification heads do not always generalise well. However, these findings are specific to the probe configurations studied, for example the transformer probe architecture used in \citet{shelmanov-etal-2025-head}, the density-based features used in \citet{vazhentsev-etal-2025-token}, or the single-token aggregation used in \citet{Levinstein_2024}. 
It therefore remains difficult to draw general conclusions about the robustness of supervised UQ probes.

UQ probing methods rely on a variety of features encompassing hidden states \cite{DBLP:conf/iclr/OrgadTGRSKB25, huang-etal-2025-confidence, he-etal-2024-llm, kossen2024semanticentropyprobesrobust, azaria-mitchell-2023-internal} and attention-based features \cite{shelmanov-etal-2025-head, ni-etal-2026-reprobe, chuang-etal-2024-lookback}. Probe architectures vary from linear probes \cite{chuang-etal-2024-lookback} and multi-layer perceptrons \cite{azaria-mitchell-2023-internal} to transformers \cite{shelmanov-etal-2025-head}, while token aggregation strategies include last-token aggregation \cite{azaria-mitchell-2023-internal, kossen2024semanticentropyprobesrobust} and mean pooling over response tokens \cite{shelmanov-etal-2025-head, vazhentsev-etal-2025-token}. 
ProbeDrift enables these design dimensions to be evaluated systematically under consistent OOD settings.

OOD evaluation varies widely between different works. This can involve training on a range of datasets with another dataset kept as a hold-out for evaluation \cite{vazhentsev-etal-2025-token, azaria-mitchell-2023-internal}, testing on synthetic data generated for different domains \cite{shelmanov-etal-2025-head}, or training with a single dataset and evaluating on the same or different tasks \cite{DBLP:conf/iclr/OrgadTGRSKB25}. As each unique OOD setting measures different levels of distributional shift, these inconsistencies make it challenging to compare conflicting claims about robustness.
ProbeDrift addresses this by providing a systematic robustness evaluation framework spanning progressively stronger distributional shifts, enabling more reliable and reproducible comparisons across methods and evaluation settings. 

Complementary concurrent work analyses layer-wise input-output behaviour in gated MLP neurons, including effects on model internal confidence \citep{gerstner2026weakening}. In contrast, we focus on the OOD robustness of supervised UQ probes.

\section{ProbeDrift: Evaluating Probe Robustness Under Distribution Shift} \label{main:existing_methods} \label{sec:robustness_of_existing_methods}

In this section we introduce the ProbeDrift evaluation framework for measuring the robustness of UQ probes across progressively stronger distributional shifts. Using ProbeDrift, we evaluate existing methods across a range of datasets and OOD settings, identifying where robustness holds and where it breaks down. 

\subsection{Experimental Setup}

ProbeDrift evaluates UQ methods across five settings of increasing difficulty. For each evaluation dataset, we test probes after training on: 1) the corresponding \textit{in-distribution} training data, 2) \textit{leave-one-out} evaluation across a range of datasets from both the same task and a different task, 3) a  \textit{single same-task} dataset, 4) a  \textit{range} of \textit{different-task} datasets, and 5) a  \textit{single different-task} dataset. 
%
% TODO: for camera-ready version, with extra page of space, go back to the bullet points below which look better:
%
%\begin{itemize}
%  \item The corresponding ID training data
%  \item A range of OOD datasets, from both the same task and another task
%  \item A single OOD dataset from the same task 
%  \item A range of datasets from a different task
%  \item A single dataset from a different task  
%\end{itemize}
%
Training data size is held constant across experiments, while the training data becomes progressively more OOD relative to the test data.

\paragraph{Datasets for Training and Evaluation.}
For evaluation, we consider four different question answering (QA) datasets, and two summarisation datasets. The QA datasets include both long-form and short-form answers. 
For QA we use SciQ \cite{welbl-etal-2017-crowdsourcing}, TriviaQA \cite{joshi-etal-2017-triviaqa}, and COQA \cite{DBLP:journals/tacl/ReddyCM19}, which contain shorter-form answers, alongside PubmedQA \cite{jin-etal-2019-pubmedqa}, which contains substantially longer generations. For summarisation we use XSum \cite{narayan-etal-2018-dont} and CNN/DailyMail \cite{DBLP:conf/nips/HermannKGEKSB15, nallapati-etal-2016-abstractive}, which also involve longer-form generations. We additionally use SamSum \cite{gliwa-etal-2019-samsum}, TruthfulQA \cite{lin-etal-2022-truthfulqa}, MMLU \cite{hendrycks2021measuring} and MedQUAD \cite{DBLP:journals/bmcbi/AbachaD19} as additional training datasets for our leave-one-out experimentation. The leave-one-out evaluation in ProbeDrift mirrors the out-of-distribution evaluation performed by \citet{vazhentsev-etal-2025-token}, which we extend to cover a wide variety of OOD settings. See Appendix \ref{appendix:train_eval_dataset_list} for further details.

\paragraph{Baselines.}
We evaluate a representative set of probing-based UQ methods spanning hidden-state probes (SAPLMA \cite{azaria-mitchell-2023-internal}), density-based probes (SATMD, SATRMD \cite{vazhentsev-etal-2025-token}), attention-based probes (Uhead \cite{shelmanov-etal-2025-head}, Lookback-Lens \cite{chuang-etal-2024-lookback}), and hybrid methods combining supervised probes with the unsupervised Maximum Sequence Probability (MSP) baseline \cite{vazhentsev-etal-2025-token, fomicheva-etal-2020-unsupervised}. We also include MSP as a simple but competitive unsupervised baseline. We report SAPLMA probes using hidden states from the middle and top layers and Uhead configurations with one or two layers. Details are provided in Appendix~\ref{appendix:uq_baselines}.
\begin{table*}[t!]
\centering
\small
\resizebox{\textwidth}{!}{%
\begin{tabular}{l c c c c c c c c c c}
\toprule
& \multicolumn{5}{c}{Llama-3.1-8B} & \multicolumn{5}{c}{Gemma-2-9B} \\
\cmidrule(lr){2-6}\cmidrule(lr){7-11}
Method & ID & LOO & 1D-SameTask & DiffTask & 1D-DiffTask
       & ID & LOO & 1D-SameTask & DiffTask & 1D-DiffTask \\
& & \multicolumn{4}{c}{\scriptsize near OOD \quad $\longrightarrow$ \quad  more OOD \quad  $\longrightarrow$ \quad  most OOD}
& & \multicolumn{4}{c}{\scriptsize near OOD \quad $\longrightarrow$ \quad  more OOD \quad  $\longrightarrow$ \quad  most OOD} \\
\midrule

\rowcolor{light-gray}\multicolumn{11}{c}{Short-form generation datasets} \\
SAPLMA (middle)    & 0.63 & \cellcolor[HTML]{DAEEF3} 0.46 & 0.26 & 0.35 & 0.32 & 0.58 & \cellcolor[HTML]{DAEEF3}0.35 & 0.11 & 0.30 & 0.21 \\
SAPLMA (top)    & 0.54 & \cellcolor[HTML]{DAEEF3} 0.26 & -0.12 & -0.14 & 0.18 & 0.56 & \cellcolor[HTML]{DAEEF3}0.34 & -0.03 & 0.06 & 0.08 \\
Uhead (1 layer)  & \cellcolor[HTML]{DAEEF3} 0.63 & 0.17 & \cellcolor[HTML]{DAEEF3}0.24 & 0.16 & 0.08  & \cellcolor[HTML]{DAEEF3}0.66 & 0.34 & \cellcolor[HTML]{DAEEF3}0.18 & -0.11 & 0.16 \\
Uhead (2 layers)  & \cellcolor[HTML]{DAEEF3} 0.63 & 0.15 & \cellcolor[HTML]{DAEEF3}0.15 & -0.27 & -0.36 & \cellcolor[HTML]{DAEEF3}0.65 & 0.33 & \cellcolor[HTML]{DAEEF3}0.20 & -0.31 & 0.12 \\
SATMD           & \cellcolor[HTML]{DAEEF3} 0.30 & 0.35 & -0.15 & 0.21 & 0.15 & \cellcolor[HTML]{DAEEF3}0.41 & 0.27 & -0.06 & 0.04 & 0.22 \\
SATRMD          & \cellcolor[HTML]{DAEEF3} 0.47 & 0.17 & -0.11 & 0.21 & 0.13 & \cellcolor[HTML]{DAEEF3}0.53 & 0.33 & 0.21 & 0.27 & 0.34 \\
Lookback-lens    & \cellcolor[HTML]{DAEEF3}0.27 & 0.00 & -0.03 & -0.13 & \cellcolor[HTML]{DAEEF3}0.19 & \cellcolor[HTML]{DAEEF3}0.34 & -0.05 & 0.09 & -0.04 & \cellcolor[HTML]{DAEEF3}0.16 \\
\midrule
SATMD-MSP        & \cellcolor[HTML]{DAEEF3} 0.44 & 0.32 & 0.06 & 0.24 & 0.23 & \cellcolor[HTML]{DAEEF3}0.62 & 0.17 & 0.15 & -0.18 & 0.41 \\
SATRMD-MSP        & \cellcolor[HTML]{DAEEF3} 0.60 & \cellcolor[HTML]{DAEEF3} 0.30 & 0.05 & 0.20 & 0.21 & \cellcolor[HTML]{DAEEF3}0.64 & \cellcolor[HTML]{DAEEF3}0.40 & 0.41 & 0.24 & 0.52 \\
HUQ-SATMD       & \cellcolor[HTML]{DAEEF3} 0.58 &  0.59 & -0.14 & 0.22 & 0.25 & \cellcolor[HTML]{DAEEF3}0.64 & 0.59 & 0.00 & 0.10 & 0.14 \\
HUQ-SATRMD    & \cellcolor[HTML]{DAEEF3} 0.61 & \cellcolor[HTML]{DAEEF3} 0.47 & -0.08 & 0.24 & 0.18 & \cellcolor[HTML]{DAEEF3}0.65 & \cellcolor[HTML]{DAEEF3}0.53 & 0.17 & 0.16 & 0.22 \\
\midrule
MSP            & 0.57 & 0.57 & 0.57 & 0.57 & 0.57 & 0.64 & 0.64 & 0.64 & 0.64 & 0.64 \\

\midrule

\rowcolor{light-gray}\multicolumn{11}{c}{Long-form generation datasets} \\
SAPLMA (middle)  & 0.32 & 0.03 & 0.07 & 0.12 & 0.04 & 0.32 & 0.01 & 0.10 & -0.05 & 0.04 \\
SAPLMA (top)      & 0.20 & 0.00 & 0.04 & 0.05 & -0.00 & 0.22 & -0.05 & 0.08 & -0.03 & 0.03 \\
Uhead (1 layer)    & \cellcolor[HTML]{DAEEF3}0.31 & -0.04 & \cellcolor[HTML]{DAEEF3}0.05 & -0.08 & -0.01 & \cellcolor[HTML]{DAEEF3}0.28 & -0.06 & \cellcolor[HTML]{DAEEF3}-0.07 & 0.05 & 0.02 \\
Uhead (2 layers)   & \cellcolor[HTML]{DAEEF3}0.27 & -0.05 & \cellcolor[HTML]{DAEEF3}0.02 & -0.03 & 0.02 & \cellcolor[HTML]{DAEEF3}0.23 & -0.09 & \cellcolor[HTML]{DAEEF3}-0.13 & -0.13 & 0.02 \\
SATMD            & \cellcolor[HTML]{DAEEF3}0.14 & -0.02 & -0.07 & 0.04 & 0.11 & \cellcolor[HTML]{DAEEF3}0.07 & 0.02 & 0.04 & -0.07 & 0.08 \\
SATRMD          & \cellcolor[HTML]{DAEEF3}0.16 & 0.03 & -0.08 & 0.05 & 0.07 & \cellcolor[HTML]{DAEEF3}0.20 & -0.15 & 0.01 & -0.13 & 0.06 \\
Lookback-lens    & \cellcolor[HTML]{DAEEF3}0.14 & 0.01 & \cellcolor[HTML]{DAEEF3}0.10 & -0.04 & \cellcolor[HTML]{DAEEF3}-0.01 & \cellcolor[HTML]{DAEEF3}0.22 & -0.04 & \cellcolor[HTML]{DAEEF3}0.10 & -0.08 & \cellcolor[HTML]{DAEEF3}-0.04 \\
\midrule
SATMD-MSP       & \cellcolor[HTML]{DAEEF3}0.18 & 0.03 & -0.03 & 0.08 & 0.12 & \cellcolor[HTML]{DAEEF3}0.18 & -0.08 & 0.02 & -0.08 & 0.09 \\
SATRMD-MSP      & \cellcolor[HTML]{DAEEF3}0.17 & \cellcolor[HTML]{DAEEF3}0.05 & -0.01 & 0.07 & 0.10 & \cellcolor[HTML]{DAEEF3}0.21 & \cellcolor[HTML]{DAEEF3}-0.15 & 0.02 & -0.08 & 0.03 \\
HUQ-SATMD       & \cellcolor[HTML]{DAEEF3}0.02 & -0.02 & -0.05 & 0.04 & 0.09 & \cellcolor[HTML]{DAEEF3}-0.01 & -0.10 & 0.06 & -0.04 & 0.08 \\
HUQ-SATRMD      & \cellcolor[HTML]{DAEEF3}0.16 & \cellcolor[HTML]{DAEEF3}0.01 & -0.08 & 0.03 & 0.04 & \cellcolor[HTML]{DAEEF3}0.20 & \cellcolor[HTML]{DAEEF3}-0.10 & 0.05 & -0.04 & -0.10 \\
\midrule
MSP             & -0.01 & -0.01 & -0.01 & -0.01 & -0.01 & -0.10 & -0.10 & -0.10 & -0.10 & -0.10 \\
\bottomrule
\end{tabular}%
}
\caption{PRR results (higher is better) across in-domain (ID) and out-of-domain (OOD) evaluation settings. Abbreviations: LOO = leave-one-out; 1D-SameTask = one dataset, same task; DiffTask = multiple datasets, different task; 1D-DiffTask = one dataset, different task. We show supervised methods, followed by hybrid methods and the MSP. Each result is an average across three different test sets. \colorbox[HTML]{DAEEF3}{Blue} highlights indicate that a corresponding evaluation was reported in the method's paper. An analysis of multi-seed variance is also provided in Appendix \ref{appendix:seed_analysis}.}
\label{tab:short_long_generation_results_two_models}
\end{table*}

\paragraph{Metrics.} Evaluating uncertainty quantification for generative tasks requires a correctness function that measures the quality of model outputs relative to reference answers. We primarily use AlignScore \cite{zha-etal-2023-alignscore}, estimating factual consistency using the entailment predictions of a RoBERTa model \cite{liu2019robertarobustlyoptimizedbert} fine-tuned on NLI and claim verification. 
AlignScore provides a continuous measure of output correctness, which we use as the target signal when evaluating uncertainty estimates. This score correlates closely with human judgements of correctness for short-form generations \cite{santilli-etal-2025-revisiting}, but without the cost of generating LLM-as-a-Judge responses. For evaluation, we further experiment with LLM-as-a-Judge, ROUGE-L, and exact match as alternative correctness functions. In Section \ref{sec:what_makes_probe_robust}, we investigate how the choice of correctness function used during training affects robustness, including training with LLM-as-a-Judge supervision.
We create feature vectors by averaging feature representations from each output token, with further aggregation strategies tested in Section \ref{main:choice_of_token_aggregation}.

Performance is evaluated using the Prediction-Rejection Ratio (PRR) \cite{vashurin-etal-2025-benchmarking, malinin-etal-2017-incorporating}, which measures how effectively an uncertainty estimate identifies unreliable outputs. PRR computes the area under the curve (AUC) for Prediction-Rejection curves, where uncertain examples are progressively rejected to measure the average correctness of the remaining examples. This score is normalised by an oracle (the best possible AUC given the generations), adjusting for a baseline that randomly predicts uncertainty:

\begin{equation}
    PRR = \frac{\text{AUC}_{\text{unc}}-\text{AUC}_{\text{rnd}}}{\text{AUC}_{\text{oracle}}-\text{AUC}_{\text{rnd}}}
  \label{eq:prr}
\end{equation}

A PRR score of 1 means that the uncertainty estimate is attaining the performance of the oracle, while a score of 0 shows chance-level performance. Scores below 0 are worse than chance-level. PRR allows for the use of a continuous correctness function such as AlignScore.

\subsection{Results}\label{sec:subsec_results}

\begin{figure*}[t]
  \centering
  \includegraphics[width=445pt]{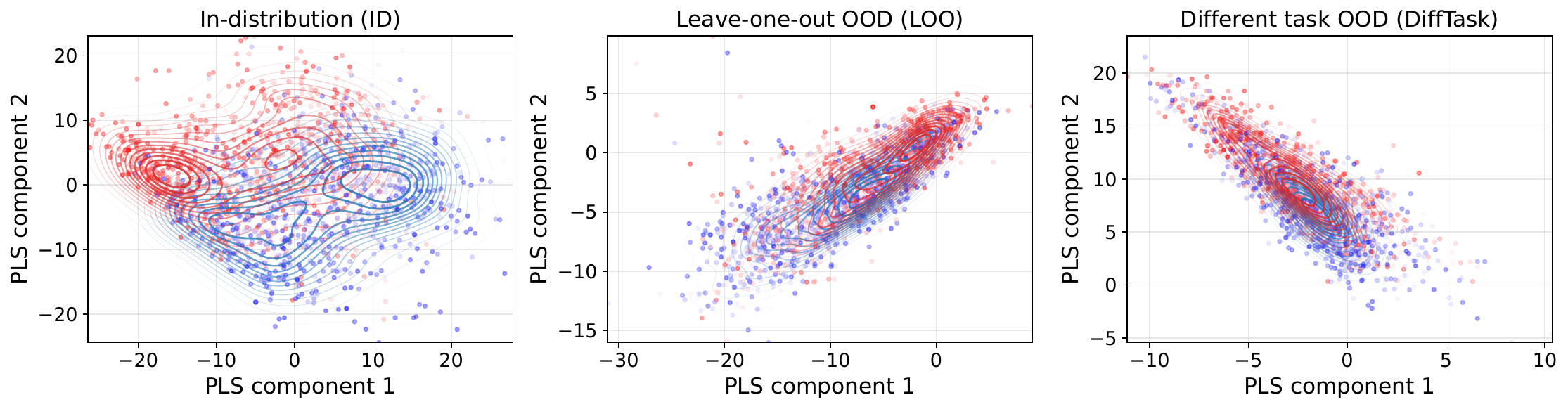}
  \caption{%Partial Least Squares (PLS) regression components for TriviaQA, where blue points represent more correct generations (higher AlignScore), and red points represent less correct generations (lower AlignScore). Kernel density estimate (KDE) contour lines show the density of generations above and below the median AlignScore.
  Visualisation of Llama 3.1-8B hidden states (layer 16) evaluated on TriviaQA, either for in-distribution, leave-one-out or different task OOD settings. Hidden states are down-projected to 2d subspaces using PLS models fitted to predict AlignScore from the training examples. As OOD-ness increases, hidden states of more correct generations (blue) and less correct generations (red) become less separable. Kernel density contour lines show the density of generations above and below the median AlignScore.
  }
  \label{fig:trivia_qa}
\end{figure*}

\begin{table*}[t!]
\centering
\small
\setlength{\tabcolsep}{4pt}
\renewcommand{\arraystretch}{0.9}
\begin{tabular}{l cc cc cc cc cc}
\toprule
& \multicolumn{2}{c}{ID} & \multicolumn{2}{c}{LOO} & \multicolumn{2}{c}{1D-SameTask} & \multicolumn{2}{c}{DiffTask} & \multicolumn{2}{c}{1D-DiffTask} \\
\cmidrule(lr){2-3}\cmidrule(lr){4-5}\cmidrule(lr){6-7}\cmidrule(lr){8-9}\cmidrule(lr){10-11}
Metric & PRR & \% & PRR & \% & PRR & \% & PRR & \% & PRR & \% \\
\midrule
\rowcolor{light-gray}\multicolumn{11}{c}{Short-form generation datasets} \\
AlignScore     & 0.63 & 100\% & 0.46 & 73\% & 0.26 & 41\% & 0.35 & 56\% & 0.32 & 51\% \\
ROUGE-L        & 0.70 & 100\% & 0.49 & 70\% & 0.30 & 43\% & 0.33 & 47\% & 0.35 & 50\% \\
Exact match       & 0.70 & 100\% & 0.48 & 69\% & 0.30 & 43\% & 0.33 & 47\% & 0.31 & 44\% \\
LLM-as-a-judge & 0.68 & 100\% & 0.55 & 81\% & 0.31  & 46\%  & 0.48 & 71\% & 0.45 & 66\% \\
\midrule
\rowcolor{light-gray}\multicolumn{11}{c}{Long-form generation datasets} \\
AlignScore     & 0.32 & 100\% & 0.03 & 9\% & 0.07 & 22\% & 0.12 & 38\% & 0.04 & 12\% \\
ROUGE-L        & 0.30 & 100\% & 0.00 & 0\% & 0.04 & 13\% & 0.11 & 37\% & 0.01 &  3\% \\
LLM-as-a-judge & 0.33 & 100\% &  0.14 & 42\%           & 0.14 & 42\% & 0.14 & 42\% & 0.08 & 24\% \\
\bottomrule
\end{tabular}
\caption{PRR results (higher is better) for SAPLMA (middle) when evaluated across four correctness functions, for Llama-3.1-8B. Each value is shown alongside its percentage of the ID score for that metric. Exact match is excluded for long-form generations as the metric is uninformative in this setting.}
\label{tab:correctness_fn_saplma_middle}
\end{table*}

UQ probes achieve strong in-distribution performance but largely fail to generalise out-of-distribution (see Table \ref{tab:short_long_generation_results_two_models}). In near-OOD settings for datasets with shorter generations, the probe design matters considerably, and the best methods reach a PRR of 0.46--0.59 for Llama (SAPLMA middle and HUQ-SATMD respectively). However, as OOD-ness increases, no method exceeds a PRR of 0.35 for Llama.
We find largely consistent trends across each dataset tested (averaged for Table \ref{tab:short_long_generation_results_two_models}), although there are some exceptions such as poor robustness on SciQ for Uhead and on TriviaQA for SATRMD in a leave-one-out setting for Llama.
Hybrid methods, which combine supervised probes with unsupervised estimates, show strong near-OOD robustness but this advantage also deteriorates under larger shifts. Notably, no probing method exceeds the unsupervised MSP baseline for any OOD setting with the exception of HUQ-SATMD in the leave-one-out setting for Llama.

When evaluating with different correctness functions, probe performance consistently degrades beyond near-OOD settings. For LLM-as-a-Judge evaluation, despite very similar findings for SAPLMA when using the last layer hidden state (see Appendix \ref{appendix:diff_correctness_functions}), we see less OOD degradation in the case of SAPLMA when using the middle layer hidden state (see Table \ref{tab:correctness_fn_saplma_middle}). 
For long-form tasks, when training with AlignScore, all methods achieve close to chance-level performance across all OOD conditions. In Section \ref{sec:what_makes_probe_robust} we show that the generalisability of probes for long-form tasks is highly dependent on the correctness function used during training, where the use of LLM-as-a-Judge for both training and evaluation can lead to substantially better reported robustness compared to using AlignScore.

We find that the strongest OOD performance is achieved by a simple probing approach (SAPLMA), and that implementation choices, such as layer selection, prove more consequential than adopting more recently proposed methods. Methods that appear robust under near-OOD evaluation also do not always maintain this advantage under broader distribution shifts (see SATMD). Finally, in Table~\ref{tab:short_long_generation_results_two_models} we show the evaluation settings closest to those used in the source papers, finding that prior evaluation tends to favour near-OOD conditions where robustness is more achievable.

\subsection{Understanding Robustness Failures}
To illustrate why probe performance deteriorates under distribution shift, particularly for long-form tasks, we visualise the hidden states using two-component Partial Least Squares (PLS) regression models that are fitted to predict the rank-normalised AlignScore. The PLS regression is used as a dimensionality reduction method, representing a strongly-regularised linear regression probe that demonstrates how uncertainty information in the hidden states becomes less accessible to probes trained on OOD data. 

Figure \ref{fig:trivia_qa} shows that correct and incorrect generations are partially separable in the in-distribution setting, indicating that correctness signals are accessible in the representation space. %\cite{marks2023geometry}. 
This structure weakens under leave-one-out training and largely collapses under different-task training.
Since linear probes can still achieve good performance when trained and evaluated on each distribution separately (Appendix \ref{appendix:hs_visualisations}), the uncertainty information itself does not disappear under distribution shift. Instead, the relationship between hidden-state representations and correctness appears to change across distributions, reducing the ability of probes trained on one distribution to generalise to another. This effect is especially pronounced for datasets with longer generations (Appendix \ref{appendix:hs_visualisations}), consistent with the weaker robustness observed in Table~\ref{tab:short_long_generation_results_two_models}.

While this section demonstrates the poor robustness for long-form generations using AlignScore, we show in Section \ref{sec:what_makes_probe_robust} that robustness is heavily dependent on the correctness metric.

\subsection{Limitations of Existing Robustness Evaluation}  \label{sec:limitations_existing_eval}

Prior work evaluates probe robustness under substantially different OOD settings, making robustness claims difficult to compare directly. Some works focus on relatively limited ``near-OOD'' settings, while others evaluate under larger distributional shifts, either within the same task or across tasks. Evaluation protocols also vary in their use of synthetic data, supervised and unsupervised baselines, and the range of distributional shifts considered.

Across the surveyed literature (with full details of the surveyed evaluation setups provided in Appendix~\ref{appendix:survey_full}), no two works evaluate on the same combination of OOD datasets. Many papers evaluate robustness under only a single type of distributional shift (54\%) and omit strong supervised or unsupervised baselines (61\% and 46\% respectively). We also find that positive robustness claims are often reported under synthetic evaluation settings, suggesting that robustness evaluation should extend beyond this setting alone.

These inconsistencies motivate the need for systematic robustness evaluation across progressively stronger distributional shifts.

\section{What makes a probe robust?} \label{sec:what_makes_probe_robust}

The results in Section \ref{sec:robustness_of_existing_methods} show that probe performance varies substantially under distribution shift. To better understand what drives these robustness differences, we systematically analyse key probe design factors, including the probe architecture, feature representations, correctness functions, and token aggregation strategies.

Guided by the findings from ProbeDrift, we perform a more detailed analysis on a representative subset of datasets, using SciQ and TriviaQA as datasets with shorter generations, and PubmedQA as a dataset with longer generations. We refer to this reduced evaluation setup as ProbeDriftLight, focusing on three representative settings: in-distribution (ID), leave-one-out OOD evaluation (LOO), and OOD evaluation after training only on datasets from a different task (DiffTask). 
We show in Appendix \ref{appendix:reduced_vs_long_probe_shift} that the ProbeDriftLight evaluation is representative of the full ProbeDrift.

We use the SAPLMA architecture in this section as it provides a simple and effective probe based on model hidden states. Results in this section are an average across both Llama-3.1-8B \cite{grattafiori2024llama3herdmodels} and Gemma-2-9B \cite{gemmateam2024gemma2improvingopen} for each dataset.\footnote{We use the same models as \citet{vazhentsev-etal-2025-token}, whose methods are our primary baseline in Section 5.}

\begin{figure}[t]
  \centering
  \includegraphics[width=\columnwidth]{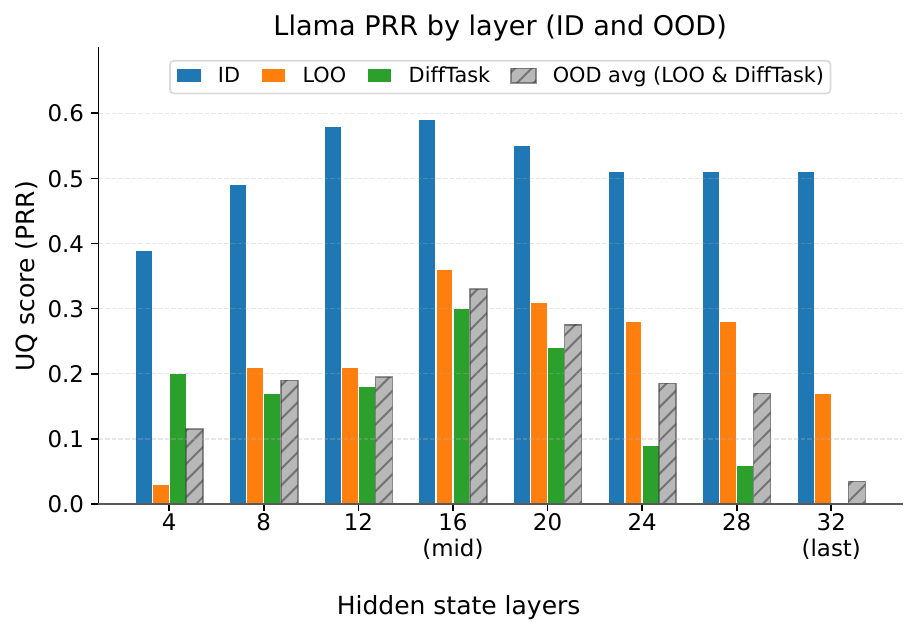}
  \caption{PRR by layer for Llama showing ID performance and two OOD settings, leave-one-out evaluation (LOO), and evaluation when training on a different task (DiffTask). OOD avg provides an average of LOO and DiffTask. Negative results floored at 0 for visualisation.}
  \label{fig:prr-by-layer}
\end{figure}

% Requires: \usepackage{booktabs}

\begin{table}[t]
\centering
\footnotesize
\setlength{\tabcolsep}{3.5pt}
\renewcommand{\arraystretch}{0.88}
\begin{tabular}{@{}lcccc@{}}
\toprule
Features used & ID & LOO & DiffTask & OOD avg \\
\midrule
Hidden state (mid. layer)     & \textbf{0.58} & \textbf{0.29} & \textbf{0.25} & \textbf{0.27} \\
Hidden state (last layer)     & 0.52 & 0.17 & -0.05 & 0.06 \\
Attention (last token)        & 0.54 & 0.28 & 0.13 & 0.21 \\
Attention (last two tokens)   & 0.55 & 0.25 & 0.15 & 0.20 \\
Lookback lens                 & 0.19 & 0.11 & -0.04 & 0.04 \\
Token probabilities           & 0.43 & 0.27 & -0.30 & -0.01 \\
\bottomrule
\end{tabular}
\vspace{-2mm}
\caption{Feature ablations, average from Llama and Gemma (PRR). OOD avg is the mean of LOO (leave-one-out) and DiffTask (different task).}
\vspace{-2mm}
\label{tab:feature_ablation_both}
\end{table}

\subsection{Choice of features}\label{main:choice_of_features}
To isolate the effect of different uncertainty signals, we fix the probe architecture and vary only the features used to train the probe. We investigate different features including attention to the previous token, attention to the last two tokens \cite{shelmanov-etal-2025-head} and the ratio of attention to the context \cite{chuang-etal-2024-lookback}. We also consider the model hidden states \cite{azaria-mitchell-2023-internal} and token probability distributions \cite{shelmanov-etal-2025-head}.

Across most features, in-distribution and leave-one-out performance is broadly similar (Table \ref{tab:feature_ablation_both}).
Differences emerge under larger distributional shifts. Hidden states from the middle layer outperform attention-based features in the DiffTask setting, while the lookback ratio and token probabilities are less effective as stand-alone signals.

Layer choice also has a strong effect on robustness (see Figure \ref{fig:prr-by-layer}).
While most layers perform comparably in-distribution, OOD performance degrades substantially at early and final layers, while middle-layer representations remain consistently more robust.
This gap widens as distributional shift increases.
Notably, using final-layer representations remains a common baseline in prior work \cite{vazhentsev-etal-2025-token, shelmanov-etal-2025-head, vazhentsev-etal-2025-unconditional}, despite its weaker robustness. In Appendix \ref{appendix:layer_wise_analysis} we further verify these findings across a range of different LLMs and evaluation metrics. 

\subsection{Choice of Token Aggregation}\label{main:choice_of_token_aggregation}

Token representations can be aggregated differently when creating the probe features. While recent work explores new variations \cite{mckenzie2025detecting, huang-etal-2025-confidence, kantamneni2025are}, the most common approach is to use the representation from the last context token \cite{DBLP:conf/nips/BurgerHN24, azaria-mitchell-2023-internal, servedio-etal-2025-hidden, kossen2024semanticentropyprobesrobust}, the final response token \cite{kossen2024semanticentropyprobesrobust}, or an average representation from the response tokens \cite{vazhentsev-etal-2025-token, shelmanov-etal-2025-head, chuang-etal-2024-lookback}. We compare the in-distribution performance and robustness of each approach, including a baseline where a response token is randomly sampled.%, and baselines using an average representation from the context tokens or an average from both the context and output tokens.

The best robustness is achieved when taking an average representation from the response tokens (see Table \ref{tab:agg_methods_both}). While similar in-distribution performance is achieved when using the last response token, we find this is less robust when tested further OOD. Interestingly, we find that randomly sampling a token from the response achieves similar robustness to using the final response token. 
In Appendix \ref{appendix:aggregation_analysis} we show that these findings are consistent across a range of LLMs and evaluation metrics.

\subsection{Training Signals from Different Correctness Functions} \label{sec:change_corretness_functions}
We experiment with changing the correctness function used during training from AlignScore to LLM-as-a-Judge, investigating the impact that this has on robustness.
For datasets with shorter generations, training with LLM-as-a-Judge leads to modest robustness improvements when evaluating with AlignScore (Table~\ref{tab:trainsignal_alignscore}). Larger improvements are observed when both training and evaluation use LLM-as-a-Judge (see Appendix~\ref{appendix:training_diff_correctness_functions}). 

For datasets with longer generations, this dependence becomes substantially stronger. Probes trained with LLM-as-a-Judge achieve near-chance performance when evaluated using AlignScore. When evaluation also uses LLM-as-a-Judge, we observe more robust probe behaviour for long-form tasks (see Appendix~\ref{appendix:training_diff_correctness_functions}). This suggests that robustness for longer generations is particularly sensitive to the correctness signal used during training and evaluation.

% Requires: \usepackage{booktabs}

\begin{table}[t]
\centering
\footnotesize
\setlength{\tabcolsep}{3.5pt}
\renewcommand{\arraystretch}{0.88}
\begin{tabular}{@{}lcccc@{}}
\toprule
Method & ID & LOO & DiffTask & OOD avg \\
\midrule

\rowcolor{light-gray}\multicolumn{5}{c}{Response tokens only} \\
Avg. of response tokens  & \textbf{0.58} & 0.29 & \textbf{0.25} & \textbf{0.27} \\
Last response token         & 0.57 & \textbf{0.30} & 0.09 & 0.20 \\
Random response token              & 0.42 & 0.24 & 0.20 & 0.22 \\
\midrule
\rowcolor{light-gray}\multicolumn{5}{c}{With context tokens} \\
Avg. of all tokens       & 0.33 & 0.09 & 0.06 & 0.07 \\
Avg. of context tokens   & 0.23 & 0.10 & 0.04 & 0.07 \\
Last context token          & 0.46 & 0.25 & 0.07 & 0.16 \\
%\midrule
%\rowcolor{light-gray}\multicolumn{5}{c}{All tokens} \\
\bottomrule
\end{tabular}
\vspace{-2mm}
\caption{Different strategies to aggregate representations across context and response tokens. Best results in bold.}
\vspace{-2mm}
\label{tab:agg_methods_both}
\end{table}

\begin{table}[t]
\centering
\small
\setlength{\tabcolsep}{4pt}
\renewcommand{\arraystretch}{0.9}
\begin{tabular}{l ccc}
\toprule
Training Signal & ID & LOO & DiffTask \\
\midrule
\rowcolor{light-gray}\multicolumn{4}{c}{Short-form} \\
AlignScore     & 0.69 & 0.52 & 0.35 \\
LLM-as-a-Judge & 0.66 & 0.56 & 0.36 \\
\midrule
\rowcolor{light-gray}\multicolumn{4}{c}{Long-form} \\
AlignScore     & 0.39 & 0.04 & 0.19 \\
LLM-as-a-Judge & -0.08 & -0.11 & 0.13 \\
\bottomrule
\end{tabular}
\caption{PRR results (higher is better) for SAPLMA (middle) evaluated with AlignScore, comparing two training signals for Llama 3.1-8B, either training with AlignScore or LLM-as-a-Judge.}
\label{tab:trainsignal_alignscore}
\end{table}

\begin{table*}
\centering
\small
\resizebox{\textwidth}{!}{%
\begin{tabular}{l c c c c c c c c c c}
\toprule
& \multicolumn{5}{c}{Llama-3.1-8B} & \multicolumn{5}{c}{Gemma-2-9B} \\
\cmidrule(lr){2-6}\cmidrule(lr){7-11}
Method & ID & LOO & 1D-SameTask & DiffTask & 1D-DiffTask
       & ID & LOO & 1D-SameTask & DiffTask & 1D-DiffTask \\
\midrule

\rowcolor{light-gray}\multicolumn{11}{c}{Short-form generation datasets} \\
\midrule

SATMD-MSP        & 0.44 & 0.32 & 0.06 & 0.24 & 0.23 & 0.62 & 0.17 & 0.15 & -0.18 & 0.41 \\
SATRMD-MSP        & 0.60 & 0.30 & 0.05 & 0.20 & 0.21 & 0.64 & 0.40 & 0.41 & 0.24 & 0.52 \\
HUQ-SATMD       & 0.58 &  0.59 & -0.14 & 0.22 & 0.25 & 0.64 & 0.59 & 0.00 & 0.10 & 0.14 \\
HUQ-SATRMD    &  0.61 & 0.47 & -0.08 & 0.24 & 0.18 & 0.65 & 0.53 & 0.17 & 0.16 & 0.22 \\
%\midrule
\midrule
MSP & 0.57 & 0.57 & \textbf{0.57} & \textbf{0.57} & \textbf{0.57} & 0.64 & \textbf{0.64} & \textbf{0.64} & \textbf{0.64} & \textbf{0.64} \\
SAPLMA (mid) & 0.63 & 0.46 & 0.26 & 0.35 & 0.32 & 0.58 & 0.35 & 0.11 & 0.30 & 0.21 \\
%\midrule
%\midrule
%MSP            & 0.57 & 0.57 & 0.57 & 0.57 & 0.57 & 0.64 & 0.64 & 0.64 & 0.64 & 0.64 \\
\midrule
HBO      & \textbf{0.66} & \textbf{0.61} & \textbf{0.57} & \textbf{0.57} & \textbf{0.57} &\textbf{0.67} & 0.61 & \textbf{0.64} & \textbf{0.64} & \textbf{0.64} \\
%HBO (adapt)  & 0.33 & 0.00 & -0.01 & 0.04 & -0.01 & \\

%\midrule
%\rowcolor{light-gray}\multicolumn{11}{c}{Long-form generation} \\
%SATMD-MSP       & 0.30 & 0.06 & 0.00 & 0.15 & 0.22 & - & - & - & - & - \\
%SATRMD-MSP      & 0.17 & 0.00 & -0.01 & 0.09 & 0.10 & - & - & - & - & - \\
%HUQ-SATMD       & - & - & - & - & - & - & - & - & - & - \\
%HUQ-SATRMD      & 0.38 & 0.30 & -0.07 & 0.16 & 0.09 & - & - & - & - & - \\
%\midrule
%MSP             & -0.01 & -0.01 & -0.01 & -0.01 & -0.01 & -0.10 & -0.10 & -0.10 & -0.10 & -0.10 \\
%\midrule
%HBO-SAPLMA      & ... & ... & ... & ... & ... & ....
\midrule
\rowcolor{light-gray}\multicolumn{11}{c}{Long-form generation datasets} \\
\midrule
SATMD-MSP       & 0.18 & 0.03 & -0.03 & 0.08 & \textbf{0.12} & 0.18 & -0.08 & 0.02 & -0.08 & \textbf{0.09} \\
SATRMD-MSP      & 0.17 & \textbf{0.05} & -0.01 & 0.07 & 0.10 & 0.21 & -0.15 & 0.02 & -0.08 & 0.03 \\
HUQ-SATMD       & 0.02 & -0.02 & -0.05 & 0.04 & 0.09 & -0.01 & -0.10 & 0.06 & \textbf{-0.04} & 0.08 \\
HUQ-SATRMD      & 0.16 & 0.01 & -0.08 & 0.03 & 0.04 & 0.20 & -0.10 & 0.05 & \textbf{-0.04} & -0.10 \\
\midrule

%\midrule
MSP             & -0.01 & -0.01 & -0.01 & -0.01 & -0.01 & -0.10 & -0.10 & -0.10 & -0.10 & -0.10 \\
SAPLMA (mid) & \textbf{0.32} & 0.03 & \textbf{0.07} & \textbf{0.12} & 0.04 & \textbf{0.32} & \textbf{0.01} & \textbf{0.10} & -0.05 & 0.04 \\

\midrule
HBO      & 0.20 & 0.02 & -0.01 & 0.03 & -0.01 & 0.03 & -0.05 & -0.10 & -0.08 & -0.10 \\
%HBO (adapt)  & 0.33 & 0.00 & -0.01 & 0.04 & -0.01 & \\
\bottomrule
\end{tabular}%
}
\caption{PRR results for HBO compared to existing hybrid methods, MSP and SAPLMA using middle layer hidden states. Results are tested in-distribution, and across the four OOD settings in ProbeDrift.}
\label{tab:hbo_short_form_results}
\end{table*}

\subsection{Probe architectures} \label{main:choice_of_architectures}

We next examine how probe architectures impact robustness. Using middle-layer hidden states as features, we compare the different probe architectures introduced in Section \ref{main:existing_methods}. We also experiment with simple modifications to the SAPLMA architecture, either adding or removing layers.

Across architectures we observe differences in in-distribution performance, but with little variation in robustness under distribution shift. The exception is linear probe architectures, which we find can be substantially less robust, although this depends on their training procedure. Overall, we find that architectural complexity and the choice of training hyper-parameters do not lead to substantially better probe robustness. 
Additional details are provided in Appendix \ref{appendix:probe_architectures}.

\section{Using ProbeDrift to Analyse and Improve Hybrid Robustness}\label{sec:introducing_hbo}

Hybrid methods have previously been proposed to combine the strong in-distribution performance of supervised probes with the broader robustness of unsupervised uncertainty estimates \cite{vazhentsev-etal-2025-token}. ProbeDrift allows us to evaluate these approaches under progressively stronger distributional shifts, revealing the weakness of existing hybrid methods beyond a near-OOD setting.

Motivated by this observation, we investigate whether robustness can be improved by adapting the contribution of these signals based on how far examples lie from the training distribution. We focus on methods introduced by \citet{vazhentsev-etal-2025-token}, which combine supervised SATRMD probes with the unsupervised MSP baseline. SATRMD-MSP incorporates MSP as an additional feature, while HUQ-SATRMD relies on MSP for in-distribution predictions, backing off to a combination of MSP and the supervised SATRMD method for OOD examples.

\subsection{A Hybrid Back-Off Strategy}

Building on our finding that supervised probe robustness degrades as distributional shift increases, we introduce a Hybrid Back-Off (HBO) strategy that progressively shifts from supervised to unsupervised uncertainty estimates. Following \citet{vazhentsev-etal-2025-token}, we estimate how OOD each test example is using Mahalanobis distance. HBO combines the robust SAPLMA probe with MSP, relying increasingly on MSP as examples move further from the training distribution. Further details are provided in Appendix~\ref{appendix:hbo_explained}.

\subsection{Results}

We find that HBO outperforms each of the hybrid baselines in almost every OOD setting for short-form generations, with larger improvements under stronger distributional shifts (see Table~\ref{tab:hbo_short_form_results}). Despite its simplicity, HBO is substantially more robust than existing approaches, illustrating how insights from comprehensive robustness evaluation can directly lead to improved robustness.
As HBO was introduced based on the strong performance of unsupervised methods for short-form generations, this method does not generalise well to long-form generations, which we leave for future work.

% An ablation experiment with a simplified version of HBO can be found in Appendix \ref{appendix:hbo_explained}.

\section{Evaluation Recommendations} \label{sec:recommendations_section}

Based on our extensive evaluation of supervised UQ probes, we identify several evaluation practices for improving the reliability of robustness claims in future work. %
%
%These recommendations reflect consistent gaps we have identified across prior work. 
%
These recommendations are informed by our experimental findings and by the gaps in existing evaluation practices highlighted in Section \ref{sec:limitations_existing_eval} and Appendix \ref{appendix:survey_full}.

We suggest future work should:

\begin{enumerate}

\itemsep-0.2em

\item Use robust baselines, including an unsupervised method using token probabilities and an MLP trained on middle-layer hidden states.
\item Show robustness for both near and far distributional shifts (e.g. both leave-one-out evaluation and training on another task).
\item Evaluate with OOD test sets used directly by previous work, rather than creating bespoke evaluation for each paper.
\item Measure robustness on both short-form and long-form generations, highlighting the room for improvement on long-form tasks.
\item Include more negative robustness results, showing the points of failure to help spur improvements in subsequent research.

\end{enumerate}

\section{Discussion and Conclusion}

Conflicting evidence about the robustness of UQ probes has made it difficult to assess whether recent methods genuinely improve OOD robustness or simply benefit from easier evaluation settings. In this work, we introduce ProbeDrift, a systematic evaluation framework for analysing robustness across progressively stronger distributional shifts.
Our results show that robustness conclusions can change substantially depending on both the evaluation regime and seemingly minor implementation choices. Methods that appear robust under limited ``near-OOD'' evaluation often degrade under broader shifts, while several commonly used evaluation settings fail to expose these weaknesses. We further find that the strongest OOD performance is achieved by a simple probing approach (SAPLMA), with implementation choices such as layer selection proving more consequential than adopting more recently proposed methods.

More broadly, our findings suggest that progress on robust uncertainty estimation is currently constrained less by probe architecture and more by the lack of systematic robustness evaluation. We find that feature selection and token aggregation decisions are important for a probe's robustness, and we also show the importance of the correctness function used in both training and evaluation, which is especially the case for long-form generations.

ProbeDrift provides a reusable framework for evaluating robustness across models, tasks, and distributional shifts, enabling more reliable comparisons between methods and helping to identify design choices that genuinely improve robustness. Beyond evaluation, we show that these insights can directly guide method development by introducing HBO, a method that adaptively shifts towards unsupervised uncertainty estimates under larger distributional shifts. We hope ProbeDrift will support future work on both developing and evaluating more robust uncertainty estimation methods.

\section*{Limitations}

In the paper we highlight how architectural advances are unlikely to lead to improvements in robustness. This is based on our experimentation testing robustness across a range of different architectures, all of which show similar levels of robustness (except for linear probes which can be less robust). While we cannot rule out the existence of a probe architecture that might improve robustness, we believe this is a less promising research direction than investigating how to create better probe inputs and training signals.

This paper involved extensive experimentation, training over 2,000 different probes (comprising more than 10k GPU hours), and therefore it will be difficult for others to fully replicate all of our experiments. We provide ProbeDriftLight to reduce the experimentation required for future OOD evaluation, which involves evaluation on three different datasets for two different OOD settings. Even with ProbeDriftLight, this additional OOD evaluation is still likely to increase the experimental burden for future work.

Due to the large computational requirement for this work, we do not perform experiments with larger LLMs. Instead, the model sizes we use are comparable with existing work \cite{vazhentsev-etal-2025-token, shelmanov-etal-2025-head, DBLP:conf/iclr/OrgadTGRSKB25, chuang-etal-2024-lookback}.

Throughout this work we train with a fixed number of training examples (1,800), keeping this consistent so different experiments can be compared on a like-for-like basis. In Appendix \ref{appendix:scale_up_train_data} we show similar findings when increasing the size of the training data (by two or four times), but we do not verify each of our findings when increasing the training data size as this would require substantially more experiments.

The OOD splits in ProbeDrift have been created to generate varying degrees of out-of-distribution, testing robustness across different OOD test sets. However, these degrees of OOD conflate several factors that are difficult to separate. For example, test sets vary in diversity, response length depends on the task and dataset, and a change in task usually brings a change in domain and task difficulty. To mitigate the effect of any single dataset’s artifacts, ProbeDrift averages results across multiple datasets for each OOD setting.

In this paper we focus on supervised probing approaches due to both their efficiency and strong performance. We compare the performance of these probing approaches to MSP, which we found to be the most competitive out of a range of unsupervised methods (see Appendix Table \ref{tab:unsupervised_baselines_llama} and Table \ref{tab:unsupervised_baselines_gemma}). We do not explore other uncertainty quantification methods, such as deep ensembles \cite{NIPS2017_9ef2ed4b} and MC Dropout \cite{DBLP:conf/icml/GalG16}.

Finally, HBO involves combining predictions from MSP and SAPLMA using their rank-normalised uncertainty scores, which assumes the whole test set is available at inference time rather than scoring examples one at a time. As PRR evaluation already requires ranking of the test set by uncertainty, the rankings required by HBO are always available.

\section*{Acknowledgements}

We are grateful to Lihu Chen,  Favour Aghaebe, Artem Shelmanov and Artem Vazhentsev for their valuable feedback. 

This work was supported by an ISPF International Collaboration Award funded by the Royal Society, as well as by the Japan Science and Technology Agency under Grant No. JST BOOST JPMJBY24F9.
We also acknowledge IT Services at The University of Sheffield for the provision of services for High Performance Computing.
AI tools (Claude and ChatGPT) were used to provide feedback on the writing in this draft and to help refactor the project code. 

\bibliography{custom}

\appendix
\clearpage

\section{Probing Baselines}\label{appendix:uq_baselines}
In this section we introduce the different supervised UQ methods that we include in Table \ref{tab:short_long_generation_results_two_models}. 

SAPLMA \cite{azaria-mitchell-2023-internal} trains a multi-layer perceptron (MLP) using the model hidden states, and we include variations of this method using either the middle or the final layer hidden states. Similar MLP probing architectures are common in existing work \cite{su-etal-2024-unsupervised, zhang2025reasoning}.

We also include baselines that train probes using attention features. Recent work by \citet{shelmanov-etal-2025-head} introduces
Uhead, a transformer-based probe using attention maps and token probabilities as features. We experiment with two Uhead architectures, reflecting the different hyper-parameter choices proposed in \citet{shelmanov-etal-2025-head}, either using one or two transformer layers for the probe. While Uhead was originally introduced to verify claim spans, we instead apply this approach to evaluate the confidence of each instance. We do this to allow for direct comparability to other methods, performing evaluation at an instance-level rather than at a claim level. Implementing Uhead at an instance level also avoids the need for claim-level training data.

Lookback-lens \cite{chuang-etal-2024-lookback} uses attention, training probes with the proportion of attention paid to the context tokens. We implement lookback-lens probes using sklearn's LinearRegression architecture rather than the paper's sklearn LogisticRegression architecture, as the LinearRegression sklearn implementation allows for training with soft target scores. 
Finally, the SATMD and SATRMD \cite{vazhentsev-etal-2025-token} baselines use density-based features, training probes using the Mahalanobis distance of hidden states from a subset of training examples, using an average distance per layer. SATRMD is a modification of SATMD using the Relative Mahalanobis distance \cite{DBLP:journals/corr/abs-2106-09022}, which normalises the Mahalanobis distance with a separate reference set.

\section{Hybrid Back-Off (HBO)}\label{appendix:hbo_explained}

HBO quantifies how far each test example lies from the training distribution using the Mahalanobis distance (MD), using this to adaptively weight the contributions of supervised and unsupervised uncertainty estimates. We describe below how an OOD ranking is computed which then determines this weighting.

Let $r$ denote the rank of a test instance MD (taken from a combination of the training MDs and the added test instance MD). We define $R$ as the normalised rank $r/(N+1)$ where $N$ is the size of the training data. We then use $R$ to decide how much to weight the rank-normalised supervised and unsupervised UQ estimates $UQ_{\text{sv}}$ and $UQ_{\text{usv}}$. We describe these weights as $W_{\text{sv}}$ and $W_{\text{usv}}$: % To do, can add word respectively at the end (if we have space in the camera-ready version
\begin{equation}
W_{\text{usv}} =
\begin{cases}
R+0.5, & \text{if } R \le 0.5,\\
1,     & \text{otherwise.}
\end{cases}
\end{equation}
\begin{align}
W_{\text{sv}} &= 1 - W_{\text{usv}}
\end{align}
\begin{align}
UQ_{\text{hyb}} = W_{\text{sv}}*UQ_{\text{sv}} + W_{\text{usv}}*UQ_{\text{usv}}
\end{align}
This approach provides an even weighting between the rank-normalised supervised and unsupervised estimates for examples closest to the training distribution, while relying more on the unsupervised estimate for OOD examples. 
Finally, based on the results from Section \ref{main:existing_methods}, we use SAPLMA as a robust probe for $UQ_{\text{sv}}$, while MSP is used for $UQ_{\text{usv}}$. We introduce HBO for short-form generations, based on the strong performance of the MSP baseline in this setting.

HBO was developed to be intentionally simple, and was designed using the SciQ dataset as a validation set where we tested two configurations: 1) the HBO configuration used in the paper, 2) a simplified configuration of HBO where $UQ_{\text{usv}} = R$ and  $UQ_{\text{sv}} = 1-R$. No additional hyper-parameter tuning was performed, and there was no further development of this method. We show the full evaluation in Tables \ref{tab:hbo_simplified_all_short}, \ref{tab:hbo_simplified_sciq} and \ref{tab:hbo_simplified_excl_sciq}, showing the performance of both HBO and the simplified configuration on SciQ, all datasets excluding SciQ, and the full ProbeDrift evaluation. While the simplified version of HBO performs well, it does not match the performance of the main configuration on short-form generation tasks.

\section{Comparing the reduced and full ProbeDrift evaluation} \label{appendix:reduced_vs_long_probe_shift}

To validate that ProbeDriftLight is representative of the full ProbeDrift evaluation, we compare results across both frameworks for each supervised method presented in Table \ref{tab:short_long_generation_results_two_models}. We report the Mean Absolute Difference (MAD) in PRR scores between the two evaluations across all methods and OOD settings (see Table~\ref{tab:MAD}). As shown in Table~\ref{tab:MAD}, differences are small across both short-form and long-form datasets, with a maximum MAD of 0.093 (long-form generations for the LOO setting), suggesting that ProbeDriftLight provides a reliable approximation of the full ProbeDrift evaluation at reduced computational cost.

\begin{table}[h]
\centering
\small
\setlength{\tabcolsep}{4pt}
\renewcommand{\arraystretch}{0.9}
\begin{tabular}{lccc}
\toprule
 & ID & LOO & DiffTask \\
\midrule
Short-form & 0.065 & 0.060 & 0.049 \\
Long-form  & 0.088 & 0.093 & 0.075 \\
\bottomrule
\end{tabular}
\caption{Mean Absolute Difference between the full ProbeDrift evaluation, and the reduced ProbeDrift evaluation comparing each of the results in Table \ref{tab:short_long_generation_results_two_models}.}
\label{tab:MAD}
\end{table}

\section{Random Seed Variance Analysis} \label{appendix:seed_analysis}

The main results in this paper are reported using a single random seed. In this section we show that while variance increases under larger distributional shifts, the overall pattern of degradation remains consistent across seeds.

We report standard deviations both for a single dataset (see Table~\ref{tab:seeds_SAMPLA_middle_sciq} and Table~\ref{tab:seeds_SAMPLA_last_sciq}) and when averaging across the three datasets used in ProbeDriftLight (see Table~\ref{tab:seeds_SAMPLA_middle_3reduced}). Variance is low in both cases, and decreases considerably when averaging across multiple datasets. For example, the standard deviation for SAPLMA (middle) drops from 0.10 in the DiffTask setting for a single dataset to 0.05 when averaged across three, supporting the reliability of the single-seed results reported in the main paper.

\begin{table}[h]
\centering
\small
\setlength{\tabcolsep}{4pt}
\renewcommand{\arraystretch}{0.9}
\begin{tabular}{lccc}
\toprule
 & ID & LOO & DiffTask \\
\midrule
1 seed        & 0.63 & 0.57 & 0.35 \\
\midrule
5 seed avg.   & 0.62 & 0.51 & 0.29 \\
\midrule
Std dev       & 0.03 & 0.07 & 0.10 \\
\bottomrule
\end{tabular}
\caption{1 seed vs.\ 5 seeds using a single dataset (SciQ) for SAPLMA (middle) for Llama.}
\label{tab:seeds_SAMPLA_middle_sciq}
\end{table}

\begin{table}[h]
\centering
\small
\setlength{\tabcolsep}{4pt}
\renewcommand{\arraystretch}{0.9}
\begin{tabular}{lccc}
\toprule
 & ID & LOO & DiffTask \\
\midrule
1 seed        & 0.61 & 0.24 & -0.33 \\
\midrule
5 seed avg.   & 0.57 & 0.08 & -0.02 \\
\midrule
Std dev       & 0.03 & 0.12 & 0.18 \\
\bottomrule
\end{tabular}
\caption{1 seed vs.\ 5 seeds using a single dataset (SciQ) for SAPLMA (top) for Llama.}
\label{tab:seeds_SAMPLA_last_sciq}
\end{table}

\begin{table}[h]
\centering
\small
\setlength{\tabcolsep}{4pt}
\renewcommand{\arraystretch}{0.9}
\begin{tabular}{lccc}
\toprule
 & ID & LOO & DiffTask \\
\midrule
1 seed        & 0.59 & 0.36 & 0.30 \\
\midrule
5 seed avg.   & 0.58 & 0.31 & 0.22 \\
\midrule
Std dev       & 0.01 & 0.03 & 0.05 \\
\bottomrule
\end{tabular}
\caption{1 seed vs.\ 5 seeds averaged across 3 datasets for SAPLMA (middle) for Llama.}
\label{tab:seeds_SAMPLA_middle_3reduced}
\end{table}

\section{Evaluation with Different Correctness Functions} \label{appendix:diff_correctness_functions}

To understand the impact of changing the correctness function used for evaluation, we provide results showing the performance of SAPLMA (middle) across four different correctness functions: AlignScore, ROUGE-L, exact match (also referred to as accuracy) and  LLM-as-a-Judge (see Table \ref{tab:correctness_fn_saplma_middle}). In Table \ref{tab:correctness_fn_saplma_top} we also show the performance of SAPLMA (top) for these same correctness functions. 

We find that AlignScore, ROUGE-L and exact match are highly correlated. For SAPLMA (top), LLM-as-a-Judge evaluation is similar to the AlignScore evaluation, however for SAPLMA (middle) we see better robustness reported using LLM-as-a-Judge (although performance still degrades further OOD in line with each of the other evaluation metrics).

To better understand these differences for SAPLMA (middle), we compare LLM-as-a-Judge evaluation to AlignScore evaluation for each specific dataset (see Table \ref{tab:llm_as_judge_results}). We find that the differences are particularly noticeable on specific datasets such as SciQ and XSum where we find substantially better robustness using LLM-as-a-Judge. 

Finally, in Table \ref{tab:correctness_functions_short_llama} we provide the full results from Table \ref{tab:short_long_generation_results_two_models} when using exact match, ROUGE-L and AlignScore for evaluation, demonstrating how closely correlated these three correctness functions are. We only provide results for short-form generations, as the exact-match correctness function is only meaningful in this setting.

\begin{table*}[t]
\centering
\small
\begin{tabular}{l c c c c}
\toprule
Dataset & Exact match & ROUGE-L & AlignScore & LLM-as-a-Judge \\
\midrule
SciQ     & 0.44 [0.30, 0.58] & 0.86 [0.72, 0.96] & 0.84 [0.68, 0.96] & 0.80 [0.68, 0.90] \\
TriviaQA & 0.76 [0.64, 0.88] & 0.92 [0.82, 1.00] & 0.84 [0.68, 0.96] & 0.94 [0.86, 1.00] \\
XSum     & 0.00 [0.00, 0.00] & 0.48 [0.24, 0.72] & 0.32 [0.04, 0.56] & 0.52 [0.30, 0.74] \\
\bottomrule
\end{tabular}
\caption{Somers' D alignment with human preferences for each correctness function, by dataset, with 95\% bootstrap confidence intervals.}
\label{tab:somers_d_human_eval}
\end{table*}

\begin{table}
\centering
\small
\resizebox{\columnwidth}{!}{%
\begin{tabular}{l c c c c c}
\toprule
& \multicolumn{5}{c}{Llama 3.1-8B}\\
\cmidrule(lr){2-6}
Eval data & ID & LOO & 1D-SameTask & DiffTask & 1D-DiffTask \\
\midrule
\rowcolor{light-gray}\multicolumn{6}{c}{PRR evaluation using AlignScore} \\
SciQ          & 0.63 & 0.57 & 0.27 & 0.35 & 0.35 \\
TriviaQA      & 0.74 & 0.48 & 0.46 & 0.36 & 0.33 \\
CoQA          & 0.52 & 0.32 & 0.03 & 0.33 & 0.27 \\
XSum          & 0.32 & 0.07 & 0.13 & 0.12 & 0.09 \\
PubmedQA      & 0.39 & 0.04 & 0.06 & 0.19 & 0.08 \\
CNN/DailyMail & 0.24 & -0.03 & 0.02 & 0.05 & -0.04 \\
\rowcolor{light-gray}\multicolumn{6}{c}{PRR evaluation using LLM-as-a-judge} \\
SciQ          & 0.69 & 0.73 & 0.24 & 0.63 & 0.61 \\
TriviaQA      & 0.79 & 0.49 & 0.52 & 0.38 & 0.38 \\
CoQA          & 0.56 & 0.41 & 0.18 & 0.44 & 0.37 \\
XSum          & 0.38 & 0.16 & 0.22 & 0.29 & 0.07 \\
PubmedQA      & 0.36 & 0.32 & 0.17 & 0.06 & 0.37 \\
CNN/DailyMail & 0.25 & -0.07 & 0.03 & 0.08 & -0.19 \\
\bottomrule
\end{tabular}%
}
\caption{PRR results (higher is better) for SAPLMA (middle layer) on Llama 3.1-8B, evaluated per dataset across in-domain (ID) and out-of-domain (OOD) settings, using AlignScore and LLM-as-a-judge as correctness functions. Abbreviations: LOO = leave-one-out OOD; 1D-SameTask = one dataset, same task (OOD); DiffTask = multiple datasets, different task (OOD); 1D-DiffTask = one dataset, different task (OOD).}
\label{tab:llm_as_judge_results}
\end{table}

\begin{table*}[t!]
\centering
\small
\setlength{\tabcolsep}{4pt}
\renewcommand{\arraystretch}{0.9}
\begin{tabular}{l cc cc cc cc cc}
\toprule
& \multicolumn{2}{c}{ID} & \multicolumn{2}{c}{LOO} & \multicolumn{2}{c}{1D-SameTask} & \multicolumn{2}{c}{DiffTask} & \multicolumn{2}{c}{1D-DiffTask} \\
\cmidrule(lr){2-3}\cmidrule(lr){4-5}\cmidrule(lr){6-7}\cmidrule(lr){8-9}\cmidrule(lr){10-11}
Metric & PRR & \% & PRR & \% & PRR & \% & PRR & \% & PRR & \% \\
& & \multicolumn{9}{c}{\scriptsize near OOD \quad $\longrightarrow$ \quad most OOD} \\
\midrule
\rowcolor{light-gray}\multicolumn{11}{c}{Short-form generation datasets} \\
AlignScore     & 0.54 & 100\% & 0.26 & 48\% & -0.12 & \textless{}0\% & -0.14 & \textless{}0\% & 0.18 & 33\% \\
ROUGE-L        & 0.60 & 100\% & 0.28 & 47\% & -0.12 & \textless{}0\% & -0.04 & \textless{}0\% & 0.13 & 22\% \\
Exact match       & 0.59 & 100\% & 0.26 & 44\% & -0.11 & \textless{}0\% &  0.01 &  2\%           & 0.10 & 17\% \\
LLM-as-a-judge & 0.60 & 100\% & 0.31 & 52\% &  -0.13  &  \textless{}0\%           & -0.24 & \textless{}0\% & 0.25 & 42\% \\
\midrule
\rowcolor{light-gray}\multicolumn{11}{c}{Long-form generation datasets} \\
AlignScore     & 0.20 & 100\% &  0.00 & 0\% & 0.04  & 20\%           &  0.05 & 25\%           & -0.00 & 0\% \\
ROUGE-L        & 0.20 & 100\% &  -0.07 & \textless{}0\% & 0.03  & 15\%           & -0.03 & \textless{}0\% &  0.00 & 0\%            \\
LLM-as-a-judge & 0.11 & 100\% &  0.04 & 36\% &  -0.06  &  \textless{}0\%           &  0.05 & 45\%           &  0.00  & 0\%            \\
\bottomrule
\end{tabular}
\caption{PRR results (higher is better) for SAPLMA (top) across four correctness functions used for evaluation (AlignScore is used for training), for Llama 3.1-8B. Each value is shown alongside its percentage of the ID score for that metric. Exact match is excluded for long-form generations as the metric is uninformative in this setting.}
\label{tab:correctness_fn_saplma_top}
\end{table*}

\begin{table}
\centering
\small
\setlength{\tabcolsep}{4pt}
\renewcommand{\arraystretch}{0.9}
\begin{tabular}{l c c c}
\toprule
Architecture & ID & LOO & DiffTask \\
\midrule

\rowcolor{light-gray}\multicolumn{4}{c}{Middle-layer hidden states} \\
SAPLMA                  & 0.58 &  0.29 & 0.25 \\
Uhead (1 layer)  & 0.52 & 0.26 & 0.21 \\
Uhead (2 layers)  & 0.50 & 0.23 & 0.13 \\
Linear regression (GD)  & 0.52 & 0.24 & \textbf{0.26} \\
Linear regression (CF)  & 0.39 & 0.12 & 0.05 \\
Linear regression (CF) + PCA & 0.50 & 0.22 & 0.16 \\
Linear regression (CF, Ridge) & 0.44 & 0.17 & 0.11 \\
\midrule

\rowcolor{light-gray}\multicolumn{4}{c}{SAPLMA variations} \\
SAPLMA (2 layer)  & 0.57 & 0.29 & 0.24 \\
SAPLMA (3 layers) & 0.57 & 0.26 & 0.24 \\
SAPLMA (5 layers) & \textbf{0.60} & \textbf{0.30} & 0.21 \\
\bottomrule
\end{tabular}
\caption{Architecture comparison on ProbeDriftLight using middle-layer hidden states (single-column version). OOD: LOO = leave-one-out; DiffTask = multiple datasets, different task. The best result (measuring PRR) in each column is in bold. Results are an average across both Llama and Gemma models. CF = closed-form; GD = gradient-descent.}
\label{tab:arch_probedriftlight_middlelayer_onecol}
\end{table}

\begin{table*}[t!]
\centering
\small
\setlength{\tabcolsep}{4pt}
\renewcommand{\arraystretch}{0.9}
\begin{tabular}{l cc cc cc cc cc}
\toprule
& \multicolumn{2}{c}{ID} & \multicolumn{2}{c}{LOO} & \multicolumn{2}{c}{1D-SameTask} & \multicolumn{2}{c}{DiffTask} & \multicolumn{2}{c}{1D-DiffTask} \\
\cmidrule(lr){2-3}\cmidrule(lr){4-5}\cmidrule(lr){6-7}\cmidrule(lr){8-9}\cmidrule(lr){10-11}
& & \multicolumn{9}{c}{\scriptsize near OOD \quad $\longrightarrow$ \quad more OOD \quad $\longrightarrow$ \quad most OOD} \\
\midrule
\rowcolor{light-gray}\multicolumn{11}{c}{AlignScore — Short-form generation datasets} \\
SAPLMA (middle)  & 0.63 & 100\% & 0.46 & 73\% & 0.26 & 41\% & 0.35 & 56\% & 0.32 & 51\% \\
SAPLMA (top)     & 0.54 & 100\% & 0.26 & 48\% & -0.12 & \textless{}0\% & -0.14 & \textless{}0\% & 0.18 & 33\% \\
Uhead (1 layer)  & 0.63 & 100\% & 0.17 & 27\% & 0.24 & 38\% & 0.16 & 25\% & 0.08 & 13\% \\
Uhead (2 layers) & 0.63 & 100\% & 0.15 & 24\% & 0.15 & 24\% & -0.27 & \textless{}0\% & -0.36 & \textless{}0\% \\
SATMD            & 0.30 & 100\% & 0.35 & 117\% & -0.15 & \textless{}0\% & 0.21 & 70\% & 0.15 & 50\% \\
SATRMD           & 0.47 & 100\% & 0.17 & 36\% & -0.11 & \textless{}0\% & 0.21 & 45\% & 0.13 & 28\% \\
Lookback-lens    & 0.27 & 100\% & 0.00 & 0\% & -0.03 & \textless{}0\% & -0.13 & \textless{}0\% & 0.19 & 70\% \\
\midrule
SATMD-MSP        & 0.44 & 100\% & 0.32 & 73\% & 0.06 & 14\% & 0.24 & 55\% & 0.23 & 52\% \\
SATRMD-MSP       & 0.60 & 100\% & 0.30 & 50\% & 0.05 & 8\% & 0.20 & 33\% & 0.21 & 35\% \\
HUQ-SATMD        & 0.58 & 100\% & 0.59 & 102\% & -0.14 & \textless{}0\% & 0.22 & 38\% & 0.25 & 43\% \\
HUQ-SATRMD       & 0.61 & 100\% & 0.47 & 77\% & -0.08 & \textless{}0\% & 0.24 & 39\% & 0.18 & 30\% \\
\midrule
\textit{Avg. supervised} & 0.50 & 100\% & 0.22 & 45\% & 0.03 & 7\% & 0.06 & 11\% & 0.10 & 20\% \\
\textit{Avg. hybrid}     & 0.56 & 100\% & 0.42 & 75\% & -0.03 & \textless{}0\% & 0.23 & 40\% & 0.22 & 39\% \\
\midrule
MSP              & 0.57 & & 0.57 & & 0.57 & & 0.57 & & 0.57 & \\
\midrule
\rowcolor{light-gray}\multicolumn{11}{c}{ROUGE-L — Short-form generation datasets} \\
SAPLMA (middle)  & 0.70 & 100\% & 0.49 & 70\% & 0.30 & 43\% & 0.33 & 47\% & 0.35 & 50\% \\
SAPLMA (top)     & 0.60 & 100\% & 0.28 & 47\% & -0.12 & \textless{}0\% & -0.04 & \textless{}0\% & 0.13 & 22\% \\
Uhead (1 layer)  & 0.73 & 100\% & 0.33 & 45\% & 0.23 & 32\% & 0.18 & 25\% & 0.01 & 1\% \\
Uhead (2 layers) & 0.74 & 100\% & 0.30 & 41\% & 0.12 & 16\% & -0.18 & \textless{}0\% & -0.12 & \textless{}0\% \\
SATMD            & 0.32 & 100\% & 0.26 & 81\% & -0.26 & \textless{}0\% & 0.27 & 84\% & 0.24 & 75\% \\
SATRMD           & 0.55 & 100\% & 0.31 & 56\% & -0.06 & \textless{}0\% & 0.26 & 47\% & 0.16 & 29\% \\
Lookback-lens    & 0.33 & 100\% & 0.13 & 39\% & -0.09 & \textless{}0\% & -0.03 & \textless{}0\% & 0.32 & 97\% \\
\midrule
SATMD-MSP        & 0.63 & 100\% & 0.04 & 6\% & -0.15 & \textless{}0\% & 0.19 & 30\% & 0.16 & 25\% \\
SATRMD-MSP       & 0.71 & 100\% & 0.29 & 41\% & 0.00 & 0\% & 0.18 & 25\% & 0.10 & 14\% \\
HUQ-SATMD        & 0.63 & 100\% & 0.60 & 95\% & -0.18 & \textless{}0\% & 0.26 & 41\% & 0.24 & 38\% \\
HUQ-SATRMD       & 0.70 & 100\% & 0.59 & 84\% & -0.02 & \textless{}0\% & 0.27 & 39\% & 0.17 & 24\% \\
\midrule
\textit{Avg. supervised} & 0.57 & 100\% & 0.30 & 53\% & 0.02 & 3\% & 0.11 & 20\% & 0.16 & 27\% \\
\textit{Avg. hybrid}     & 0.67 & 100\% & 0.38 & 57\% & -0.09 & \textless{}0\% & 0.23 & 34\% & 0.17 & 25\% \\
\midrule
MSP              & 0.63 & & 0.63 & & 0.63 & & 0.63 & & 0.63 & \\
\midrule
\rowcolor{light-gray}\multicolumn{11}{c}{Accuracy — Short-form generation datasets} \\
SAPLMA (middle)  & 0.70 & 100\% & 0.48 & 69\% & 0.30 & 43\% & 0.33 & 47\% & 0.31 & 44\% \\
SAPLMA (top)     & 0.59 & 100\% & 0.26 & 44\% & -0.11 & \textless{}0\% & 0.01 & 2\% & 0.10 & 17\% \\
Uhead (1 layer)  & 0.74 & 100\% & 0.32 & 43\% & 0.23 & 31\% & 0.15 & 20\% & 0.03 & 4\% \\
Uhead (2 layers) & 0.75 & 100\% & 0.31 & 41\% & 0.10 & 13\% & -0.21 & \textless{}0\% & -0.06 & \textless{}0\% \\
SATMD            & 0.27 & 100\% & 0.26 & 96\% & -0.17 & \textless{}0\% & 0.25 & 93\% & 0.18 & 67\% \\
SATRMD           & 0.56 & 100\% & 0.34 & 61\% & -0.09 & \textless{}0\% & 0.22 & 39\% & 0.09 & 16\% \\
Lookback-lens    & 0.29 & 100\% & 0.10 & 34\% & -0.03 & \textless{}0\% & -0.03 & \textless{}0\% & 0.40 & 138\% \\
\midrule
SATMD-MSP        & 0.63 & 100\% & 0.10 & 16\% & -0.09 & \textless{}0\% & 0.15 & 24\% & 0.10 & 16\% \\
SATRMD-MSP       & 0.71 & 100\% & 0.32 & 45\% & -0.02 & \textless{}0\% & 0.11 & 15\% & 0.03 & 4\% \\
HUQ-SATMD        & 0.69 & 100\% & 0.66 & 96\% & -0.13 & \textless{}0\% & 0.23 & 33\% & 0.17 & 25\% \\
HUQ-SATRMD       & 0.75 & 100\% & 0.64 & 85\% & -0.04 & \textless{}0\% & 0.21 & 28\% & 0.08 & 11\% \\
\midrule
\textit{Avg. supervised} & 0.56 & 100\% & 0.30 & 53\% & 0.03 & 6\% & 0.10 & 18\% & 0.15 & 27\% \\
\textit{Avg. hybrid}     & 0.69 & 100\% & 0.43 & 62\% & -0.07 & \textless{}0\% & 0.17 & 25\% & 0.10 & 14\% \\
\midrule
MSP              & 0.69 & & 0.69 & & 0.69 & & 0.69 & & 0.69 & \\
\bottomrule
\end{tabular}
\caption{PRR results (higher is better) for Llama 3.1-8B across three correctness functions (AlignScore, ROUGE-L, and exact match accuracy), for short-form generation datasets only. Each value is shown alongside its percentage of the ID score for that method. Average rows show the mean PRR across supervised and hybrid methods respectively, with percentages computed relative to the averaged ID score.}
\label{tab:correctness_functions_short_llama}
\end{table*}

\section{Training with Different Correctness Functions} \label{appendix:training_diff_correctness_functions}

Table~\ref{tab:trainsignal_saplma} provides results comparing AlignScore and LLM-as-a-Judge as training signals for SAPLMA (middle). These results include evaluation across multiple different correctness functions. 

For short-form generations, training with LLM-as-a-Judge improves robustness, with the largest improvements observed after evaluating with LLM-as-a-Judge itself. However, we also see improvements over training with AlignScore when AlignScore is used for evaluation. This suggests that LLM-as-a-Judge provides a more generalisable training signal.

\begin{table*}[h]
  \centering
  \begin{tabular}{l rrr rr}
  \toprule
   & \multicolumn{3}{c}{\textbf{Scoring judgements}} & \multicolumn{2}{c}{\textbf{Binary judgements}} \\
  \cmidrule(lr){2-4} \cmidrule(lr){5-6}
  Dataset & 0 & (0,\,1) & 1 & 0 & 1 \\
  \midrule
  COQA     & 9.6\%  & 8.5\%  & 82.0\% & 14.1\% & 85.9\% \\
  PubmedQA   &  5.2\% & 17.4\% & 77.4\% & 10.0\% & 90.0\% \\
  SciQ     &  4.2\% &  4.5\% & 91.3\% &  7.0\% & 93.0\% \\
  TriviaQA & 35.2\% &  3.9\% & 61.0\% & 37.5\% & 62.5\% \\
  \bottomrule
  \end{tabular}
  \caption{Score distributions for scoring LLM-as-a-Judge judgements vs binary judgements. We show results for our QA datasets, as binary judgements are less suitable for summarisation tasks.}
  \label{tab:llm_judge_score_distributions}
\end{table*}

\begin{table}[t]
\centering
\small
\setlength{\tabcolsep}{4pt}
\renewcommand{\arraystretch}{0.9}
\begin{tabular}{l cc cc cc}
\toprule
& \multicolumn{2}{c}{ID} & \multicolumn{2}{c}{LOO} & \multicolumn{2}{c}{DiffTask} \\
\cmidrule(lr){2-3}\cmidrule(lr){4-5}\cmidrule(lr){6-7}
Eval Metric & PRR & \% & PRR & \% & PRR & \% \\
\midrule
\rowcolor{light-gray}\multicolumn{7}{c}{\textit{Short-form — Trained with AlignScore}} \\
AlignScore     & 0.69 & 100\% & 0.52 & 77\% & 0.35 & 52\% \\
ROUGE-L        & 0.78 & 100\% & 0.57 & 73\% & 0.32 & 40\% \\
Accuracy       & 0.77 & 100\% & 0.57 & 75\% & 0.32 & 41\% \\
LLM-judge  & 0.74 & 100\% & 0.61 & 82\% & 0.51 & 68\% \\
\midrule
\rowcolor{light-gray}\multicolumn{7}{c}{\textit{Short-form — Trained with LLM-as-a-judge}} \\
AlignScore     & 0.66 & 100\% & 0.56 & 86\% & 0.36 & 55\% \\
ROUGE-L        & 0.62 & 100\% & 0.56 & 90\% & 0.39 & 62\% \\
Accuracy       & 0.57 & 100\% & 0.53 & 92\% & 0.35 & 62\% \\
LLM-judge  & 0.86 & 100\% & 0.72 & 84\% & 0.55 & 64\% \\
\midrule
\rowcolor{light-gray}\multicolumn{7}{c}{\textit{Long-form — Trained with AlignScore}} \\
AlignScore     & 0.39 & 100\% & 0.04 &  10\% & 0.19 & 49\% \\
ROUGE-L        & 0.40 & 100\% & -0.17 &  \textless{}0\% & 0.09 & 22\% \\
LLM-judge  & 0.36 & 100\% & 0.32 & 89\% & 0.06 & 17\% \\
\midrule
\rowcolor{light-gray}\multicolumn{7}{c}{\textit{Long-form — Trained with LLM-as-a-judge}} \\
AlignScore     & -0.08 & -  & -0.11 & -  & 0.13 & - \\
ROUGE-L        & -0.35 & -  & -0.41 & -  & -0.02 & - \\
LLM-judge & 0.69 & 100\% & 0.31 & 45\% & 0.46 & 67\% \\
\bottomrule
\end{tabular}
\caption{PRR results (higher is better) for SAPLMA (middle) trained with different training signals, evaluated with four correctness functions, for Llama 3.1-8B with ProbeDriftLight. Accuracy is excluded as an evaluation metric for long-form generations (as this is not applicable in these cases). Each value is shown alongside its percentage of the ID score for that metric.}
\label{tab:trainsignal_saplma}
\end{table}

\section{Human Evaluation of Correctness Functions} \label{appendix:human_eval_correctness_functions}

We perform a small human evaluation to check the appropriateness of the different correctness functions used in this work. This involves an author of the paper assessing pairs of test examples, measuring whether human preferences align with preferences from each correctness function. We measure the alignment to human preferences using Somers’ D, where we treat the human preferences as the reference variable. As a result, we do not consider examples where human judgement has no preference (81\% of pairs for SciQ), leaving a sample of 50 pairs per dataset. The human evaluator only had access to the prompt, reference answer, and the LLM output when giving their preference, and did not see any of the correctness function scores. As a result of this small sample size, this evaluation can be seen as a sanity check on our choice of correctness functions.

Table \ref{tab:somers_d_human_eval} suggests that ROUGE-L, AlignScore and LLM-as-a-Judge are all appropriate correctness functions for short-form generations, however determining correctness for long-form generations is substantially more challenging. Table \ref{tab:human_eval_pairs} shows the pairs used for each dataset, including the number of pairs where human annotation expressed no preference.

\begin{table*}[t]
\centering
\small
\begin{tabular}{l c c c c}
\toprule
Dataset & Pairs seen (incl. ties) & \# Ties & Sample size & Tie rate \\
\midrule
SciQ     & 261 & 211 & 50 & 80.8\% \\
TriviaQA & 109 & 59  & 50 & 54.1\% \\
XSum     & 91  & 41  & 50 & 45.1\% \\
\bottomrule
\end{tabular}
\caption{Number of pairs sampled for the human evaluation, before and after excluding pairs where human annotation expressed no preference.}
\label{tab:human_eval_pairs}
\end{table*}

\section{Evaluation with AlignScore for Long-Form Generations} \label{appendix:limitations_longform_alignscore}

In Table \ref{tab:long_form_ablation}, we show each of the different design choices using the PubmedQA long-form dataset in ProbeDriftLight after training with AlignScore. We demonstrate consistent OOD failures across each of these different design choices, suggesting the limitations of using AlignScore in this setting.

\section{Layer Choice Analysis} \label{appendix:layer_wise_analysis}

Table~\ref{tab:layer_ablation_per_model} verifies that the findings from Section~\ref{main:choice_of_features} hold consistently across a range of LLMs and evaluation metrics. Across all six models tested, in-distribution performance is broadly similar across layers, while OOD performance degrades substantially at early and final layers. Middle-layer representations (including the middle layer itself and for nearby layers) are consistently more robust across AlignScore, ROUGE-L, and exact match accuracy, with the exception of Qwen-2.5-3B-IT which performs better for later layers.
% Requires (in preamble):
% \usepackage{booktabs}
% \usepackage[table]{xcolor}
% \definecolor{light-gray}{gray}{0.94}

\begin{table}[!h]
\centering
\small
\setlength{\tabcolsep}{4pt}
\renewcommand{\arraystretch}{0.9}
\begin{tabular}{l c c}
\toprule
 & Train $\rho$ & Test $\rho$  \\
\midrule

\rowcolor{light-gray}\multicolumn{3}{c}{SciQ} \\
\midrule

ID & 0.7421 & 0.5163 \\
LOO & 0.5432 & 0.4633 \\
DiffTask & 0.5528 & 0.2518 \\
\midrule

\rowcolor{light-gray}\multicolumn{3}{c}{TriviaQA} \\
\midrule

ID & 0.5988 & 0.5385 \\
LOO & 0.5733 & 0.2601 \\
DiffTask & 0.5528 & 0.2738 \\
\midrule

\rowcolor{light-gray}\multicolumn{3}{c}{PubmedQA} \\
\midrule

ID & 0.3754 & 0.3526  \\
LOO & 0.6221 & -0.0674  \\
DiffTask & 0.5528 & -0.0903 \\

\bottomrule
\end{tabular}
\caption{Spearman's rank correlation coefficient for PLS regression models for both the training and test data. }
\label{tab:spearmans_rank}
\end{table}

\section{Aggregation Choice Analysis} \label{appendix:aggregation_analysis}

Table~\ref{tab:aggregation_llama} verifies that the findings from Section~\ref{main:choice_of_token_aggregation} hold across different evaluation metrics. Across AlignScore, ROUGE-L, exact match, and LLM-as-a-Judge, averaging over response tokens achieves the best PRR for most OOD settings. Context-based aggregation strategies consistently underperform response-token strategies across all metrics.

Additionally, we compare using an average from each of the response tokens to results using only the last response token, testing this across a variety of different LLMs (see Table \ref{tab:token_aggregation_alignscore}). These were the two strategies with the highest in-distribution performance in Table 
\ref{tab:agg_methods_both}. This analysis shows how using the average aggregation is more robust for all of the models further OOD (in the DiffTask setting).

\section{Hidden State Visualisations} \label{appendix:hs_visualisations}
\begin{figure*}[t]
  \centering
  \includegraphics[width=445pt]{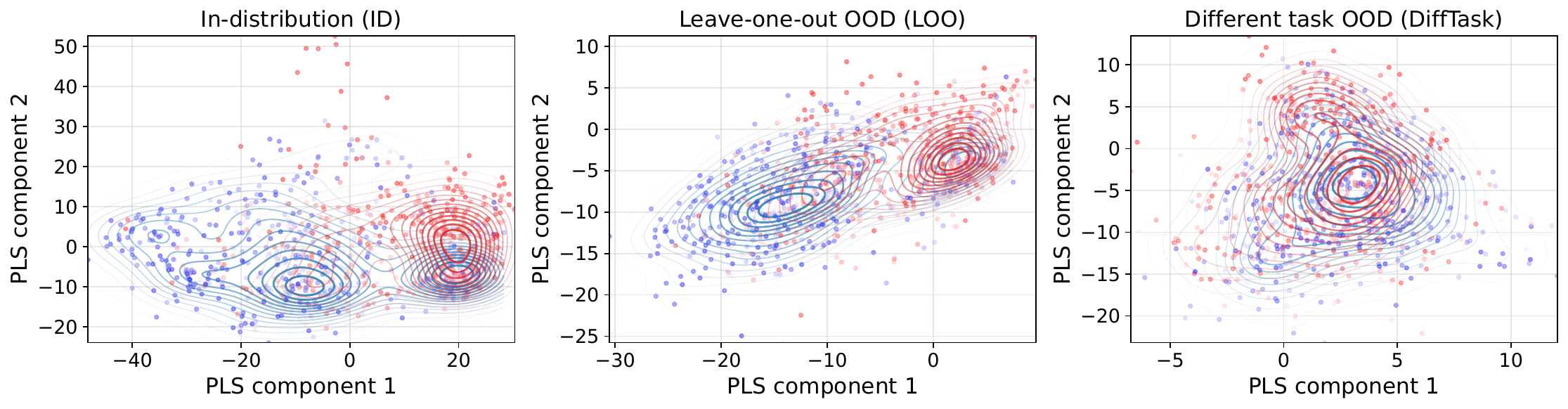}
  \caption{
    Visualisation of Llama 3.1-8B hidden states (layer 16) evaluated on SciQ, either for in-distribution, leave-one-out or different task OOD settings. Hidden states are down-projected to 2d subspaces using PLS models fitted to predict AlignScore from the training examples. As OOD-ness increases, hidden states of more correct generations (blue) and less correct generations (red) become less separable. Kernel density contour lines show the density of generations above and below the median AlignScore.}
  \label{fig:sciq}
\end{figure*}

This section shows the Partial Least Squares regression visualisations for both SciQ and PubmedQA. As expected, for SciQ the graphs show how less correct generations are less easily distinguishable from more correct generations in further OOD settings (see Figure \ref{fig:sciq}).

\begin{figure*}[t]
  \centering
  \includegraphics[width=445pt]{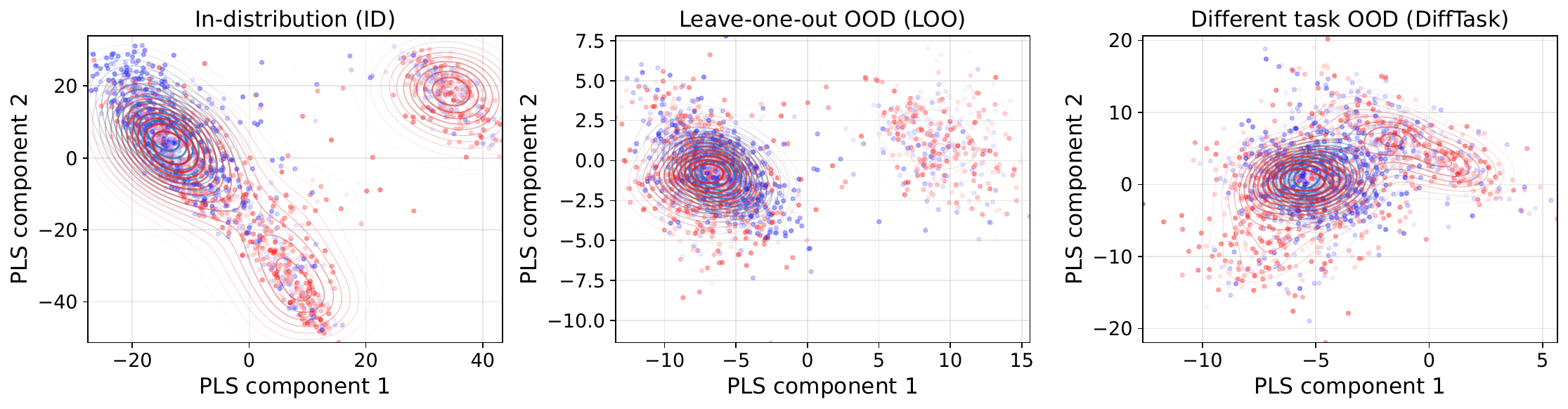}
  \caption{
  Visualisation of Llama 3.1-8B hidden states (layer 16) evaluated on PubmedQA, either for in-distribution, leave-one-out or different task OOD settings. Hidden states are down-projected to 2d subspaces using PLS models fitted to predict AlignScore from the training examples. As OOD-ness increases, hidden states of more correct generations (blue) and less correct generations (red) become less separable. Kernel density contour lines show the density of generations above and below the median AlignScore.}
  \label{fig:pubmed}
\end{figure*}

For PubmedQA (see Figure \ref{fig:pubmed}) there are two distinct groups of examples, which upon closer inspection represent examples where the model provides a yes/no answer without further elaboration (the smaller group, which have low AlignScores), and where the model provides an elaboration alongside its answer (the larger group). We find that separating these two types of answers is insufficient for good performance. Within the larger group of examples, for the in-distribution graph we can see that the correct and less correct examples are somewhat distinguishable, although not to the same extent as for SciQ and TriviaQA. This is explained by the lower PRR observed on PubmedQA (0.39 for SAPLMA, see Table \ref{tab:hallucination_probes}), compared to SciQ and TriviaQA (0.63 and 0.74 respectively). Similar to the other datasets, these examples become less distinguishable in further OOD settings. 

Table \ref{tab:spearmans_rank} shows the Spearman's rank correlation coefficients for both the training and test data for the PLS regression models. 

\section{Full Analysis of Past Experimentation}
\label{appendix:survey_full}

Table \ref{tab:survey_ood_experimentation_all_papers} provides the full list of papers analysed for their OOD experimentation and claims. We include probing papers from 2024 onwards, by which point SAPLMA \cite{azaria-mitchell-2023-internal} was an established approach.
For each paper, we investigate the different distributional shifts tested, considering evaluation in a leave-one-out setting (which we describe as near-OOD), evaluation after training on a single same-task dataset, and evaluation after training only on different-task datasets. We also highlight work that evaluates OOD performance on at least 3 different test sets, and papers that do not rely entirely on synthetic evaluation. Finally, we consider which papers include a strong unsupervised baseline (using the model soft probabilities, such as MSP or Perplexity), or a strong supervised baseline (training a probe with a middle layer hidden state, which for Table \ref{tab:survey_ood_experimentation_all_papers} we define as any hidden state that is not from the first or last layer). 

\section{Probe Architectures}
\label{appendix:probe_architectures}

We experiment with a range of different probe architectures when using the middle layer hidden state from the model (which the best performing SAPLMA baseline used). Similar to the other results in Section \ref{sec:what_makes_probe_robust}, we consider an average from Gemma-2-9B and Llama 3.1-8B across SciQ, TriviaQA and PubmedQA. We experiment with the SAPLMA architecture, a four layer MLP with hidden layers of size 256, 128 and 64, with a single output logit on the final layer. We also experiment with the two Uhead architectures used by \citet{shelmanov-etal-2025-head} after their hyper-parameter tuning. The single layer Uhead architecture involves a single transformer layer with 16 attention heads, while the 2 layer Uhead has 4 attention heads. Both architectures have 768 dimensional hidden states.
We additionally experiment with a single linear layer \cite{chuang-etal-2024-lookback}, and a linear layer after applying PCA \cite{vazhentsev-etal-2025-token}. Based on the success of SAPLMA, we also experiment with removing either one or two linear layers (removing those with the largest dimensionality), or introducing an additional layer with a 512 dimensional hidden state (resulting in 5 layers with 512, 256, 128 and 64 dimensional hidden states respectively).

We find that all the probes tested (with the exception of linear probes) achieve very similar levels of robustness (see Table \ref{tab:arch_probedriftlight_middlelayer_onecol}). When evaluating in a leave-one-out OOD setting, all other architectures are within 0.07 percentage points. We conclude that the choice of architecture makes less difference to the probe's robustness (although interestingly we see larger differences for in-distribution performance). For linear probes, their robustness depends on the training procedure, with probes trained with gradient descent achieving similar robustness to other probing architectures, while closed-form methods (including when trained with PCA or ridge regression regularisation) have worse robustness.

Our results also suggest that additional hyper-parameter tuning for the probes is unlikely to further improve robustness. We find similar performance across each type of probe with their different parameters (for example comparing the two Uhead configurations). Moreover, as hyper-parameter tuning is also likely to be conducted on an in-distribution validation set, we argue that this is unlikely to lead to substantial improvements in robustness.

\begin{table}[t]
\centering
\small
\setlength{\tabcolsep}{4pt}
\renewcommand{\arraystretch}{0.9}
\begin{tabular}{l ccc}
\toprule
Training set size & ID & LOO & DiffTask \\
\midrule
\rowcolor{light-gray}\multicolumn{4}{c}{SAPLMA (middle)} \\
Normal (1x) & 0.59 & 0.36 & 0.30 \\
x2          & 0.59 & 0.37 & 0.17 \\
x4          & 0.62 & 0.40 & 0.32 \\
\midrule
\rowcolor{light-gray}\multicolumn{4}{c}{SAPLMA (top)} \\
Normal (1x) & 0.51 & 0.17 & -0.10 \\
x2          & 0.52 & 0.06 & 0.02 \\
x4          & 0.56 & 0.22 & 0.09 \\
\bottomrule
\end{tabular}
\caption{PRR results (higher is better) for SAPLMA (middle) and SAPLMA (top), comparing training set size (Normal, 2x, and 4x the standard size) for Llama-3.1-8B with ProbeDriftLight.}
\label{tab:training_size_scaling}
\end{table}

\section{LLM-as-a-Judge Setup} \label{appendix:llm_as_a_judge}

For our LLM-as-a-Judge we allow for either a correct judgement (a score of 1), an incorrect judgement (a score of 0), or a partially correct judgement with a score between 0 and 1. We do this to avoid class imbalances which cause higher variance with the PRR metric, which considers the extent to which more correct instances are more confident than less correct instances. We show the difference in distribution between both approaches in Table \ref{tab:llm_judge_score_distributions}, where the binary-judgement LLM-as-a-Judge setup mirrors the judge in \citet{DBLP:conf/iclr/OrgadTGRSKB25}.

\begin{figure*}[t]
  \centering
  \includegraphics[width=460pt]{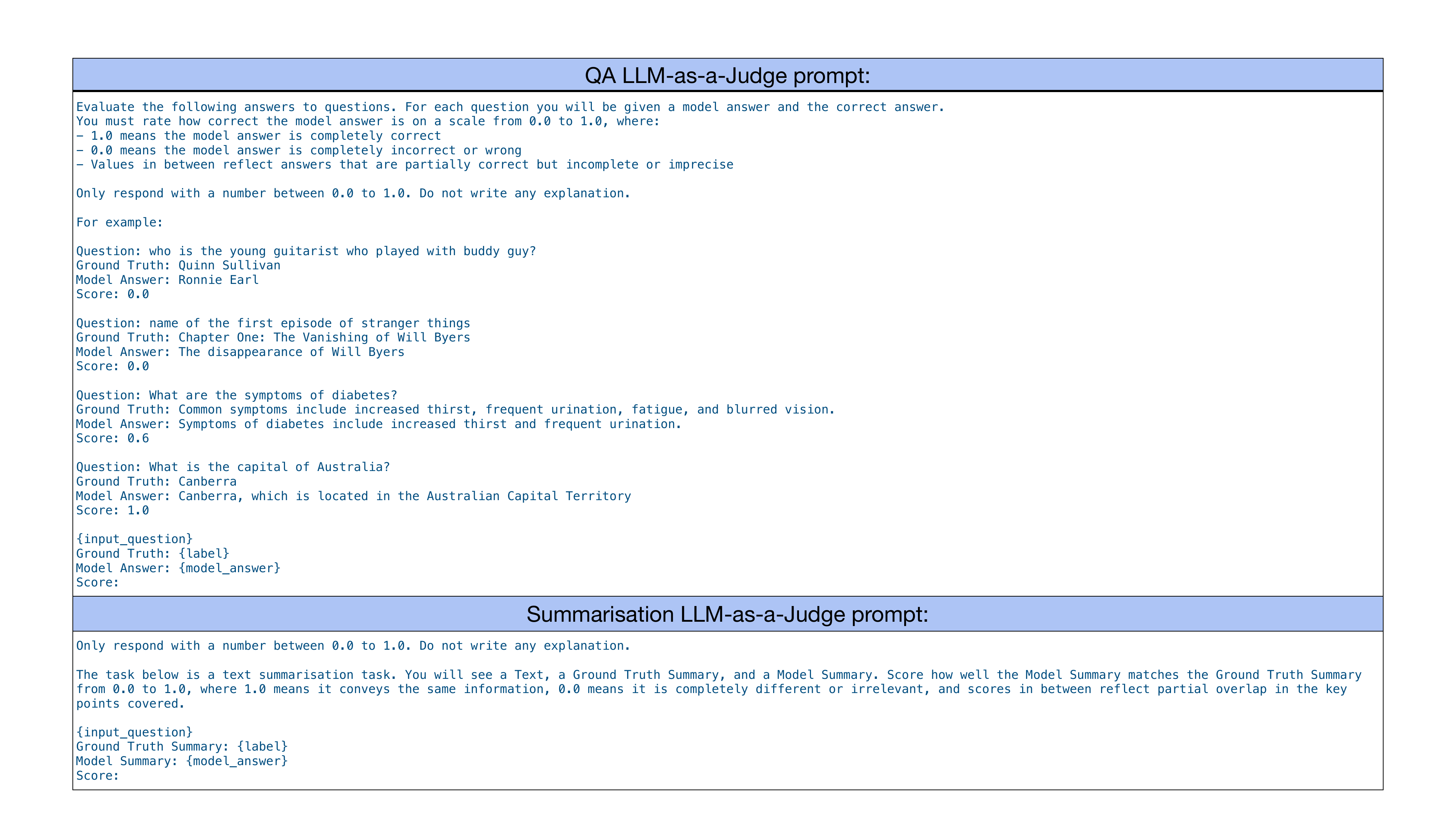}
  \caption{
    LLM prompts used for question answering and summarisation tasks.}
  \label{fig:llm_judge_prompts}
\end{figure*}

We use GPT-5 (gpt-5-2025-08-07) \cite{singh2026openaigpt5card} for the LLM-as-a-Judge. The soft scoring approach used in the paper uses the prompts in Figure \ref{fig:llm_judge_prompts}. Human preferences are used to assess the LLM-as-a-Judge, comparing the LLM-as-a-Judge to other correctness functions (see Appendix \ref{appendix:human_eval_correctness_functions}):

\section{The performance of unsupervised baselines} \label{appendix:unsupervised_baselines}

Table \ref{tab:unsupervised_baselines_llama} and Table \ref{tab:unsupervised_baselines_gemma} show the performance of a range of unsupervised methods for Llama-3.1 and Gemma-2 respectively. We find that MSP (Maximum Sequence Probability) performs best for short-form generations on average across both models, and we therefore consider this as our main unsupervised baseline. For long-form generations all methods perform poorly (with close to random chance performance). 

The unsupervised baselines include MSP, Perplexity, Lexical Similarity \cite{fomicheva-etal-2020-unsupervised}, the DegMat, Eccentricity and EigValLaplacian methods proposed by \citet{DBLP:journals/tmlr/LinT024}, Semantic Entropy \cite{kuhn2023semantic} and SAR \cite{duan-etal-2024-shifting}.

\section{Scaling up Training Data Size} \label{appendix:scale_up_train_data}

Throughout this work, we fix the number of training examples to allow a like-for-like comparison between experiments, training probes with 1,800 training instances. In Table \ref{tab:training_size_scaling} we also experiment with training with more data, using either two or four times the amount of training data. 

When increasing the size of the training data, there are small improvements in in-distribution performance. There are also some improvements in out-of-distribution performance, although these are less consistent. 
We find similar behaviour when increasing the size of the training data, with probes on the middle layer hidden states being consistently more robust than probes trained on the last layer hidden states.

\section{Training and Evaluation Datasets used OOD} \label{appendix:train_eval_dataset_list}

Table \ref{tab:full_list_train_eval_data} provides a comprehensive list of each of the datasets used for training for each OOD setting and dataset used for evaluation. In each case, we fix the number of training instances to 1,800. We evaluate on at most 2,000 test instances (following \citet{vazhentsev-etal-2025-token}).

\section{Additional Baseline for Long-Form Answers} \label{appendix:addition_long_form_baseline}

We additionally consider the probing method proposed by \citet{obeso2025realtimedetectionhallucinatedentities} specifically for long-form generations. This approach involves additional annotation for each training instance to create hallucination labels for different entities. Therefore, we use the probes provided by \citet{obeso2025realtimedetectionhallucinatedentities} which are trained on a sample of 10,000 instances from LongFact \cite{DBLP:conf/nips/WeiYSLH0TPLHDL24}, LongFact++ and TriviaQA. The training process involves using a LoRA adapter to change the behaviour of the original baseline model in addition to training a linear probe. For evaluation, we consider each test instance as a single entity.

% Requires (in preamble):
% \usepackage{booktabs}
% \usepackage[table]{xcolor}
% \definecolor{light-gray}{gray}{0.94}

\begin{table}
\centering
\small
\setlength{\tabcolsep}{4pt}
\renewcommand{\arraystretch}{0.9}
\begin{tabular}{l c}
\toprule
Dataset & Avg. words  \\
\midrule

\rowcolor{light-gray}\multicolumn{2}{c}{Short-form generation datasets} \\
\midrule

SciQ & 1.5 \\
TriviaQA & 2.3  \\
COQA & 2.5 \\
\midrule

Avg. Short form & 2.1 \\
\midrule

\rowcolor{light-gray}\multicolumn{2}{c}{Long-form generation datasets} \\
\midrule

XSum & 20.9  \\
PubmedQA & 37.1\\
CNN & 54.4  \\
\midrule

Avg. Long form & 37.5 \\

\bottomrule
\end{tabular}
\caption{Average dataset answer lengths.}
\label{tab:response_len}
\end{table}

% Preamble (make sure these are present)

\definecolor{light-gray}{gray}{0.94}

\begin{table*}[t]
\centering
\scriptsize % smaller than \footnotesize
\setlength{\tabcolsep}{2pt}
\renewcommand{\arraystretch}{0.85}

\newlength{\colone}
\newlength{\coleq}
\setlength{\colone}{0.10\textwidth}
\setlength{\coleq}{\dimexpr(\textwidth-\colone-10\tabcolsep)/5\relax}

\begin{tabular}{@{}%
>{\raggedright\arraybackslash}p{\colone}
*{5}{>{\raggedright\arraybackslash}p{\coleq}}
@{}}
\toprule
Evaluation data & ID & LOO & 1D-SameTask & DiffTask & 1D-DiffTask \\
\midrule

\rowcolor{light-gray}\multicolumn{6}{c}{\textbf{QA evaluation data}} \\
\midrule

SciQ \newline (1000) & SciQ \newline (1800) &
SamSum (200), \newline
XSum (200), \newline
CNN-Dailymail (200), \newline
COQA (200), \newline
TriviaQA (200), \newline
MMLU (200), \newline
PubmedQA (200), \newline
MedQUAD (200), \newline
TruthfulQA (200)
& MedQUAD \newline (1800) &
SamSum (600), \newline
XSum (600), \newline
CNN-Dailymail (600)
& SamSum \newline (1800) \\
\addlinespace[2pt]
\midrule
TriviaQA \newline (2000) & TriviaQA \newline (1800) &
SamSum (200), \newline
XSum (200), \newline
CNN-Dailymail (200), \newline
COQA (200), \newline
SciQ (200), \newline
MMLU (200), \newline
PubmedQA (200), \newline
MedQUAD (200), \newline
TruthfulQA (200)
& MedQUAD \newline (1800) &
SamSum (600), \newline
XSum (600), \newline
CNN-Dailymail (600)
& SamSum \newline (1800) \\
\addlinespace[2pt]
\midrule
PubmedQA \newline (2000) & PubmedQA \newline (1800) &
SamSum (200), \newline
XSum (200), \newline
CNN-Dailymail (200), \newline
COQA (200), \newline
SciQ (200), \newline
MMLU (200), \newline
TriviaQA (200), \newline
MedQUAD (200), \newline
TruthfulQA (200)
& MedQUAD \newline (1800) &
SamSum (600), \newline
XSum (600), \newline
CNN-Dailymail (600)
& SamSum \newline (1800) \\
\addlinespace[2pt]
\midrule

COQA \newline (2000) & COQA \newline (1800) &
SamSum (200), \newline
XSum (200), \newline
CNN-Dailymail (200), \newline
PubmedQA (200), \newline
SciQ (200), \newline
MMLU (200), \newline
TriviaQA (200), \newline
MedQUAD (200), \newline
TruthfulQA (200)
& MedQUAD \newline (1800) &
SamSum (600), \newline
XSum (600), \newline
CNN-Dailymail (600)
& SamSum \newline (1800) \\
\addlinespace[2pt]
\midrule

%MMLU \newline (2000) & \textcolor{red}{MMLU \newline (1461)} &
%SamSum (200), \newline
%XSum (200), \newline
%CNN-Dailymail (200), \newline
%PubMed (200), \newline
%SciQ (200), \newline
%MedQUAD (200), \newline
%TriviaQA (200), \newline
%COQA (200), \newline
%TruthfulQA (200)
%& PubMed \newline (1800) &
%SamSum (600), \newline
%XSum (600), \newline
%CNN-Dailymail (600)
%& SamSum \newline (1800) \\
%\midrule

%MedQUAD \newline (2000) & \textcolor{red}{MedQUAD \newline (1000)} &
%SamSum (200), \newline
%XSum (200), \newline
%CNN-Dailymail (200), \newline
%PubMed (200), \newline
%SciQ (200), \newline
%MMLU (200), \newline
%TriviaQA (200), \newline
%COQA (200), \newline
%TruthfulQA (200)
%& PubMed \newline (1800) &
%SamSum (600), \newline
%XSum (600), \newline
%CNN-Dailymail (600)
%& SamSum \newline (1800) \\
%\midrule
\rowcolor{light-gray}\multicolumn{6}{c}{\textbf{Summarisation evaluation data}} \\
\midrule

XSum \newline (2000) & XSum \newline (1800) &
SamSum (200), \newline
SciQ (200), \newline
CNN-Dailymail (200), \newline
COQA (200), \newline
TriviaQA (200), \newline
MMLU (200), \newline
PubmedQA (200), \newline
MedQUAD (200), \newline
TruthfulQA (200)
& SamSum \newline (1800) &
SciQ (300), \newline
MMLU (300), \newline
TriviaQA (300), \newline
COQA (300), \newline
TruthfulQA (300), \newline
MedQUAD (300)
& MedQUAD \newline (1800) \\
\addlinespace[2pt]
\midrule

CNN-Dailymail \newline (2000) & CNN-Dailymail \newline (1800) &
SamSum (200), \newline
SciQ (200), \newline
XSum (200), \newline
COQA (200), \newline
TriviaQA (200), \newline
MMLU (200), \newline
PubmedQA (200), \newline
MedQUAD (200), \newline
TruthfulQA (200)
& SamSum \newline (1800) &
SciQ (300), \newline
MMLU (300), \newline
TriviaQA (300), \newline
COQA (300), \newline
TruthfulQA (300), \newline
MedQUAD (300)
& MedQUAD \newline (1800) \\

\bottomrule
\end{tabular}

\caption{For each dataset used for evaluation, the different datasets used for training are provided for each OOD setting. In each case, the number of training examples are provided next to each dataset (the total is always 1,800 for each experiment). The total amount of evaluation data used for each dataset is also provided in the first column (we sample at most 2000 test instances).}
\label{tab:full_list_train_eval_data}
\end{table*}

\begin{table*}
\centering
\small
\resizebox{\textwidth}{!}{ %
\begin{tabular}{l c c c c c c c c}
\toprule
& \multicolumn{3}{c}{Short-form} & & \multicolumn{3}{c}{Long-form} \\
\cmidrule(lr){2-4}\cmidrule(lr){6-8}
Method & SciQ & COQA & TriviaQA & \textbf{Avg. Short} & XSum & CNN & PubmedQA & \textbf{Avg. Long} \\
\midrule

\rowcolor{light-gray}\multicolumn{9}{c}{Llama 3.1-8B results} \\

MSP  & 0.58 & 0.45 & 0.69 & 0.57 & -0.13 & 0.10 & -0.01 & -0.01 \\
Perplexity      & 0.20 & 0.45 & 0.69 & 0.45 & -0.11 & 0.24 & -0.04 & 0.03 \\
DegMat NLI Score Ent       & 0.55 & 0.43 & 0.72 & 0.57 & 0.06 & 0.16 & 0.07 & \textbf{0.10} \\
Eccentricity NLI Score Ent     & 0.52 & 0.44 & 0.65 & 0.54 & 0.02 & 0.05 & 0.01 & 0.03    \\
Eig ValLaplacian NLI Score Ent   & 0.52 & 0.47 & 0.71 & 0.57 & 0.06 & 0.16 & 0.07 & \textbf{0.10} \\
Lexical Similarity ROUGE-L  & 0.49 & 0.42 & 0.71 & 0.54 & 0.02 & 0.09 & 0.05 & 0.05  \\
SAR  & 0.45 & 0.45 & 0.76 & 0.55 & 0.01 & 0.13 & 0.02 & 0.05  \\
Semantic Entropy   & 0.52 & 0.44 & 0.56 & 0.51 & 0.02 & 0.08 & 0.02 & 0.04    \\
SentenceSAR  & 0.55 & 0.46 & 0.73 & \textbf{0.58} & -0.04 & 0.06 & -0.03 & -0.00  \\
%\midrule

\bottomrule
\end{tabular}%
}
\caption{Performance of unsupervised methods using Llama-3.1-8B on our chosen datasets, using PRR with AlignScore for evaluation.}
\label{tab:unsupervised_baselines_llama}
\end{table*}

\begin{table*}
\centering
\small
\resizebox{\textwidth}{!}{ %
\begin{tabular}{l c c c c c c c c}
\toprule
& \multicolumn{3}{c}{Short-form} & & \multicolumn{3}{c}{Long-form} \\
\cmidrule(lr){2-4}\cmidrule(lr){6-8}
Method & SciQ & COQA & TriviaQA & \textbf{Avg. Short} & XSum & CNN & PubmedQA & \textbf{Avg. Long} \\
\midrule

\rowcolor{light-gray}\multicolumn{9}{c}{Gemma-2-9B results} \\

MSP  & 0.61 & 0.53 & 0.77 & \textbf{0.64} & -0.14 & 0.05 & -0.21 & -0.10\\
Perplexity & 0.18 & 0.52 & 0.73 & 0.48 & -0.19 & 0.01 & 0.20 & 0.01 \\
DegMat NLI Score Ent &  0.51 & 0.42 & 0.74 & 0.56 & 0.09 & 0.10 & 0.02 & \textbf{0.07} \\
Eccentricity NLI Score Ent  & 0.50 & 0.48 & 0.72 & 0.57 & 0.02 & 0.06 & -0.00 & 0.03   \\
Eig ValLaplacian NLI Score Ent & 0.48 & 0.45 & 0.75 & 0.56 & 0.09 & 0.10 & 0.03 & \textbf{0.07} \\
Lexical Similarity ROUGE-L  & 0.47 & 0.44 & 0.74 & 0.55 & 0.06 & 0.10 & 0.05 & \textbf{0.07} \\
SAR  & 0.44 & 0.48 & 0.75 & 0.56 & 0.03 & 0.11 & 0.03 & 0.06 \\
Semantic Entropy   & 0.49 & 0.49 & 0.60 & 0.53 & 0.07 & 0.08 & 0.02 & 0.06 \\
SentenceSAR  & 0.58 & 0.53 & 0.75 & 0.62 & -0.00 & 0.06 & -0.06 & -0.00 \\
%\midrule

\bottomrule
\end{tabular}%
}
\caption{Performance of unsupervised methods using Gemma-2-9B on our chosen datasets, using PRR with AlignScore for evaluation.}
\label{tab:unsupervised_baselines_gemma}
\end{table*}

\begin{table*}
\centering
\small
\resizebox{\textwidth}{!}{ %
\begin{tabular}{l c c c c c c c c}
\toprule
& \multicolumn{3}{c}{Short-form} & & \multicolumn{3}{c}{Long-form} \\
\cmidrule(lr){2-4}\cmidrule(lr){6-8}
Method & SciQ & COQA & TriviaQA & \textbf{Avg. Short} & XSum & CNN & PubmedQA & \textbf{Avg. Long} \\
\midrule

\rowcolor{light-gray}\multicolumn{9}{c}{Llama 3.1-8B results} \\
\midrule
SAPLMA (ID)  & 0.63 & 0.52 & 0.74 & 0.63 & 0.32 & 0.24 & 0.39 & 0.32 \\
SAPLMA (LOO) & 0.57 & 0.32 & 0.48 & 0.46 & 0.07 & -0.03 & 0.04 & 0.03  \\
SAPLMA (1D-SameTask) & 0.27 & 0.03 & 0.46 & 0.26 & 0.13 & 0.02 & 0.06 & 0.07 \\
SAPLMA (DiffTask)    & 0.35 & 0.33 & 0.36 & 0.35 & 0.12 & 0.05 & 0.19 & 0.12  \\
SAPLMA (1D-DiffTask) & 0.35 & 0.27 & 0.33 & 0.32 & 0.09 & -0.04 & 0.08 & 0.04 \\
\midrule
\citet{obeso2025realtimedetectionhallucinatedentities}  & 0.50 & 0.26 & 0.37 & 0.38 & 0.01 & -0.01 & -0.05 & -0.02 \\
%\midrule
\bottomrule
\end{tabular}%
}
\caption{We compare the robustness of the method proposed by \citet{obeso2025realtimedetectionhallucinatedentities}, showing no better robustness on long-form tasks compared to the SAPLMA probe (using the middle-layer hidden states).}
\label{tab:hallucination_probes}
\end{table*}

As we present in Table \ref{tab:hallucination_probes}, despite the much larger training sample, the additional entity-level annotation, and the use of the LoRA adapter, we do not find more robust behaviour from these probes. Similar to the other probes considered, performance on the long-form datasets is similar to random chance performance.

\section{Answer Length by Dataset} \label{appendix:task_response_length}

Table \ref{tab:response_len} shows the answer lengths (number of words) for each dataset, showing which datasets are categorised as long-form and short-form based on the length of their answers. The short-form generations have 2.1 words on average per answer, compared to 37.5 for the long-form generations.

\begin{table*}[t]
\centering
\footnotesize
\setlength{\tabcolsep}{4pt}
\renewcommand{\arraystretch}{0.65}
\begin{tabular}{l ccc ccc ccc}
\toprule
& \multicolumn{3}{c}{AlignScore} & \multicolumn{3}{c}{ROUGE-L} & \multicolumn{3}{c}{Accuracy} \\
\cmidrule(lr){2-4}\cmidrule(lr){5-7}\cmidrule(lr){8-10}
Layer & ID & LOO & DiffTask & ID & LOO & DiffTask & ID & LOO & DiffTask \\
\midrule
\rowcolor{light-gray}\multicolumn{10}{c}{Llama-3.1-8B} \\
middle-12 & 0.39 & 0.03 & 0.20 & 0.47 & 0.07 & \textbf{0.25} & 0.50 & 0.39 & 0.34 \\
middle-8  & 0.49 & 0.21 & 0.17 & 0.57 & 0.21 & 0.15 & 0.59 & 0.35 & \textbf{0.38} \\
middle-4  & 0.58 & 0.21 & 0.18 & 0.63 & 0.26 & 0.14 & 0.64 & 0.46 & 0.21 \\
middle    & \textbf{0.59} & \textbf{0.36} & \textbf{0.30} & \textbf{0.65} & \textbf{0.32} & 0.24 & \textbf{0.68} & \textbf{0.55} & \textbf{0.38} \\
middle+4  & 0.55 & 0.31 & 0.24 & 0.59 & 0.27 & 0.19 & 0.62 & 0.50 & 0.37 \\
middle+8  & 0.51 & 0.28 & 0.09 & 0.56 & 0.26 & 0.12 & 0.60 & 0.46 & 0.24 \\
middle+12 & 0.51 & 0.28 & 0.06 & 0.57 & 0.23 & 0.10 & 0.61 & 0.47 & 0.26 \\
last      & 0.51 & 0.17 & -0.10 & 0.57 & 0.12 & -0.08 & 0.61 & 0.36 & 0.12 \\
\midrule
\rowcolor{light-gray}\multicolumn{10}{c}{Llama-3.1-8B-IT} \\
middle-12 & 0.45 & 0.10 & -0.02 & 0.34 & 0.11 & 0.02 & 0.47 & 0.31 & 0.20 \\
middle-8  & 0.55 & 0.09 & 0.05 & 0.45 & 0.11 & 0.05 & 0.54 & 0.30 & 0.26 \\
middle-4  & 0.62 & \textbf{0.31} & 0.06 & 0.55 & 0.25 & 0.08 & 0.62 & 0.44 & 0.28 \\
middle    & \textbf{0.64} & 0.29 & \textbf{0.11} & \textbf{0.56} & \textbf{0.28} & \textbf{0.13} & \textbf{0.64} & \textbf{0.48} & \textbf{0.33} \\
middle+4  & 0.62 & \textbf{0.31} & 0.06 & 0.51 & \textbf{0.28} & 0.05 & 0.63 & 0.46 & 0.24 \\
middle+8  & 0.59 & 0.24 & 0.01 & 0.50 & 0.23 & 0.06 & 0.60 & 0.43 & 0.26 \\
middle+12 & 0.59 & 0.22 & 0.02 & 0.48 & 0.22 & 0.07 & 0.60 & 0.41 & 0.27 \\
last      & 0.55 & 0.25 & 0.01 & 0.46 & 0.26 & 0.05 & 0.58 & 0.43 & 0.26 \\
\midrule
\rowcolor{light-gray}\multicolumn{10}{c}{Gemma-2-9B} \\
middle-16 & 0.35 & 0.06 & 0.11 & 0.40 & 0.01 & 0.07 & 0.46 & 0.33 & 0.23 \\
middle-12 & 0.44 & 0.02 & 0.07 & 0.54 & -0.07 & -0.03 & 0.50 & 0.28 & 0.26 \\
middle-8  & 0.50 & 0.15 & 0.15 & 0.56 & 0.07 & -0.01 & 0.57 & 0.40 & 0.23 \\
middle-4  & 0.56 & 0.17 & 0.25 & 0.64 & 0.12 & \textbf{0.28} & 0.60 & 0.47 & 0.29 \\
middle    & 0.58 & 0.22 & 0.20 & 0.64 & 0.15 & 0.09 & 0.62 & 0.48 & 0.38 \\
middle+4  & 0.59 & \textbf{0.29} & \textbf{0.27} & \textbf{0.69} & \textbf{0.22} & 0.17 & \textbf{0.64} & \textbf{0.53} & \textbf{0.44} \\
middle+8  & \textbf{0.60} & 0.26 & 0.21 & \textbf{0.69} & 0.18 & 0.16 & 0.63 & 0.52 & 0.37 \\
middle+12 & 0.55 & 0.26 & 0.17 & 0.63 & 0.19 & 0.12 & 0.60 & 0.52 & 0.32 \\
middle+16 & 0.53 & 0.24 & 0.17 & 0.64 & 0.16 & 0.12 & 0.60 & 0.49 & 0.31 \\
last      & 0.54 & 0.17 & 0.01 & 0.65 & 0.12 & 0.03 & 0.62 & 0.44 & 0.14 \\
\midrule
\rowcolor{light-gray}\multicolumn{10}{c}{OLMo-2-7B} \\
middle-12 & 0.33 & 0.10 & 0.08 & 0.32 & 0.08 & 0.07 & 0.28 & 0.15 & 0.13 \\
middle-8  & 0.42 & 0.15 & 0.10 & 0.42 & 0.13 & 0.08 & 0.41 & 0.18 & 0.16 \\
middle-4  & 0.52 & 0.28 & 0.18 & 0.49 & 0.23 & 0.19 & 0.48 & 0.24 & 0.27 \\
middle    & \textbf{0.58} & \textbf{0.42} & \textbf{0.37} & \textbf{0.56} & \textbf{0.43} & \textbf{0.37} & \textbf{0.55} & \textbf{0.47} & 0.31 \\
middle+4  & 0.54 & 0.35 & 0.33 & 0.53 & 0.36 & 0.34 & 0.51 & 0.38 & \textbf{0.37} \\
middle+8  & 0.52 & 0.34 & 0.26 & 0.49 & 0.30 & 0.22 & 0.49 & 0.36 & 0.23 \\
middle+12 & 0.51 & 0.33 & 0.14 & 0.46 & 0.29 & 0.13 & 0.48 & 0.30 & 0.13 \\
last      & 0.48 & 0.31 & 0.16 & 0.44 & 0.29 & 0.16 & 0.47 & 0.35 & 0.14 \\
\midrule
\rowcolor{light-gray}\multicolumn{10}{c}{Falcon-7B} \\
middle-12 & 0.46 & 0.03 & 0.16 & 0.54 & -0.10 & 0.28 & 0.52 & 0.27 & 0.26 \\
middle-8  & 0.50 & 0.10 & 0.16 & 0.60 & -0.08 & 0.25 & 0.56 & 0.29 & 0.26 \\
middle-4  & \textbf{0.56} & 0.23 & 0.17 & 0.61 & 0.09 & 0.05 & 0.60 & 0.44 & \textbf{0.37} \\
middle    & \textbf{0.56} & \textbf{0.25} & 0.20 & 0.65 & 0.11 & 0.15 & 0.61 & 0.44 & 0.30 \\
middle+4  & 0.54 & 0.20 & 0.18 & 0.64 & 0.10 & 0.16 & \textbf{0.62} & 0.44 & 0.25 \\
middle+8  & 0.54 & \textbf{0.25} & \textbf{0.26} & \textbf{0.66} & \textbf{0.13} & \textbf{0.31} & \textbf{0.62} & \textbf{0.45} & 0.31 \\
middle+12 & 0.52 & 0.21 & 0.22 & 0.63 & 0.10 & 0.17 & 0.59 & \textbf{0.45} & 0.28 \\
last      & 0.51 & 0.16 & 0.10 & 0.61 & 0.05 & -0.01 & 0.57 & 0.39 & 0.22 \\
\midrule
\rowcolor{light-gray}\multicolumn{10}{c}{Qwen-2.5-3B-IT} \\
middle-12 & 0.46 & 0.06 & -0.00 & 0.55 & 0.04 & -0.04 & 0.49 & 0.32 & 0.24 \\
middle-8  & 0.48 & 0.09 & 0.01 & 0.56 & 0.17 & -0.19 & 0.49 & 0.28 & 0.14 \\
middle-4  & 0.54 & 0.14 & 0.08 & 0.61 & 0.19 & -0.01 & 0.54 & 0.36 & 0.20 \\
middle    & 0.61 & 0.09 & 0.03 & 0.70 & 0.11 & -0.18 & 0.62 & 0.36 & 0.12 \\
middle+4  & 0.64 & 0.21 & 0.13 & 0.72 & 0.20 & \textbf{0.12} & 0.65 & 0.45 & 0.25 \\
middle+8  & \textbf{0.67} & \textbf{0.34} & \textbf{0.20} & \textbf{0.76} & \textbf{0.37} & -0.01 & \textbf{0.68} & \textbf{0.55} & \textbf{0.29} \\
middle+12 & 0.65 & 0.23 & 0.07 & 0.74 & 0.23 & -0.06 & 0.65 & 0.45 & 0.19 \\
last      & 0.59 & 0.22 & 0.11 & 0.68 & 0.17 & -0.02 & 0.60 & 0.44 & 0.20 \\
\midrule
\rowcolor{light-gray}\multicolumn{10}{c}{Average (all models)} \\
middle-12 & 0.42 & 0.06 & 0.08 & 0.46 & 0.02 & 0.09 & 0.46 & 0.29 & 0.24 \\
middle-8  & 0.49 & 0.13 & 0.11 & 0.53 & 0.10 & 0.05 & 0.53 & 0.30 & 0.24 \\
middle-4  & 0.56 & 0.22 & 0.15 & 0.59 & 0.19 & 0.12 & 0.58 & 0.40 & 0.27 \\
middle    & \textbf{0.59} & 0.27 & \textbf{0.20} & \textbf{0.63} & 0.23 & 0.13 & \textbf{0.62} & \textbf{0.46} & 0.30 \\
middle+4  & 0.58 & 0.28 & \textbf{0.20} & 0.61 & 0.24 & \textbf{0.17} & 0.61 & \textbf{0.46} & \textbf{0.32} \\
middle+8  & 0.57 & \textbf{0.29} & 0.17 & 0.61 & \textbf{0.25} & 0.15 & 0.61 & \textbf{0.46} & 0.28 \\
middle+12 & 0.55 & 0.26 & 0.11 & 0.58 & 0.21 & 0.09 & 0.59 & 0.43 & 0.24 \\
last      & 0.53 & 0.21 & 0.05 & 0.57 & 0.17 & 0.02 & 0.57 & 0.40 & 0.18 \\
\bottomrule
\end{tabular}
\caption{PRR results (higher is better) by representation layer across three correctness functions and six LLMs, evaluated on ProbeDriftLight. Bold indicates the best value per column within each model section. We consider the following models: Llama-3.1-8B \cite{grattafiori2024llama3herdmodels}, Llama 3.1-8B-IT, Gemma-2-9B \cite{gemmateam2024gemma2improvingopen}, OLMo-2-7B \cite{olmo20252olmo2furious}, Falcon-7B \cite{almazrouei2023falconseriesopenlanguage} and Qwen-2.5-3B-IT \cite{qwen2025qwen25technicalreport}.}
\label{tab:layer_ablation_per_model}
\end{table*}

\begin{table*}
\centering
\small
\begin{tabular}{ll ccc}
\toprule
Model & Method & ID & LOO & DiffTask \\
\midrule
Llama-3.1-8B & Avg response tokens & \textbf{0.59} & \textbf{0.36} & \textbf{0.30} \\
             & Last response token  & 0.58 & \textbf{0.36} & 0.16 \\
\midrule
Llama-3.1-8B-IT & Avg response tokens & \textbf{0.64} & \textbf{0.29} & \textbf{0.11} \\
                & Last response token  & 0.61 & 0.21 & -0.06 \\
\midrule
Gemma-2-9B & Avg response tokens & \textbf{0.58} & 0.22 & \textbf{0.20} \\
           & Last response token  & 0.57 & \textbf{0.25} & 0.02 \\
\midrule
OLMo-2-7B & Avg response tokens & \textbf{0.58} & \textbf{0.42} & \textbf{0.37} \\
          & Last response token  & 0.53 & 0.40 & 0.28 \\
\midrule
Falcon-7B & Avg response tokens & \textbf{0.56} & 0.25 & \textbf{0.20} \\
          & Last response token  & \textbf{0.56} & \textbf{0.26} & 0.19 \\
\midrule
Qwen-2.5-3B-IT & Avg response tokens & \textbf{0.61} & 0.09 & \textbf{0.03} \\
               & Last response token  & 0.56 & \textbf{0.17} & -0.09 \\
\midrule
Average & Avg response tokens & \textbf{0.59} & \textbf{0.27} & \textbf{0.20} \\
        & Last response token  & 0.57 & \textbf{0.27} & 0.08 \\
\bottomrule
\end{tabular}
\caption{PRR results for SAPLMA using AlignScore, comparing average vs last response token aggregation, averaged across SciQ, TriviaQA, and PubmedQA.  We consider the following models: Llama-3.1-8B \cite{grattafiori2024llama3herdmodels}, Llama 3.1-8B-IT, Gemma-2-9B \cite{gemmateam2024gemma2improvingopen}, OLMo-2-7B \cite{olmo20252olmo2furious}, Falcon-7B \cite{almazrouei2023falconseriesopenlanguage} and Qwen-2.5-3B-IT \cite{qwen2025qwen25technicalreport}.}
\label{tab:token_aggregation_alignscore}
\end{table*}

\begin{table}[t]
\centering
\small
\setlength{\tabcolsep}{4pt}
\renewcommand{\arraystretch}{0.9}
\begin{tabular}{l ccc}
\toprule
Aggregation Method & ID & LOO & DiffTask \\
\midrule
\rowcolor{light-gray}\multicolumn{4}{c}{AlignScore} \\
Average of response tokens & 0.59 & 0.36 & 0.30 \\
Last response token        & 0.58 & 0.36 & 0.16 \\
Random response token      & 0.41 & 0.27 & 0.24 \\
Average of all tokens      & 0.38 & 0.05 & 0.17 \\
Average of context tokens  & 0.32 & 0.07 & 0.16 \\
Last context token         & 0.46 & 0.25 & 0.13 \\
\midrule
\rowcolor{light-gray}\multicolumn{4}{c}{ROUGE-L} \\
Average of response tokens & 0.65 & 0.32 & 0.24 \\
Last response token        & 0.63 & 0.35 & 0.21 \\
Random response token      & 0.46 & 0.22 & 0.16 \\
Average of all tokens      & 0.47 & 0.02 & 0.04 \\
Average of context tokens  & 0.29 & 0.03 & 0.05 \\
Last context token         & 0.42 & 0.19 & 0.02 \\
\midrule
\rowcolor{light-gray}\multicolumn{4}{c}{Accuracy} \\
Average of response tokens & 0.68 & 0.55 & 0.38 \\
Last response token        & 0.66 & 0.51 & 0.33 \\
Random response token      & 0.54 & 0.45 & 0.32 \\
Average of all tokens      & 0.51 & 0.22 & 0.19 \\
Average of context tokens  & 0.37 & 0.20 & 0.20 \\
Last context token         & 0.52 & 0.36 & 0.18 \\
\midrule
\rowcolor{light-gray}\multicolumn{4}{c}{LLM-as-a-judge} \\
Average of response tokens & 0.61 & 0.51 & 0.35 \\
Last response token        & 0.56 & 0.50 & 0.17 \\
Random response token      & 0.45 & 0.35 & 0.28 \\
Average of all tokens      & 0.40 & 0.19 & 0.36 \\
Average of context tokens  & 0.36 & 0.15 & 0.33 \\
Last context token         & 0.50 & 0.43 & 0.23 \\
\bottomrule
\end{tabular}
\caption{PRR results (higher is better) for SAPLMA (middle) across token aggregation strategies and four evaluation metrics, for Llama-3.1-8B. Results are averaged across SciQ, TriviaQA, and PubmedQA.}
\label{tab:aggregation_llama}
\end{table}

\begin{table}[t]
\centering
\small
\setlength{\tabcolsep}{4pt}
\renewcommand{\arraystretch}{0.9}
\begin{tabular}{l c c c}
\toprule
& ID & LOO & DiffTask \\
\midrule
\rowcolor{light-gray}\multicolumn{4}{c}{Representation layer} \\
middle-12 & 0.35 & -0.16 & 0.02 \\
middle-8  & 0.36 & -0.05 & 0.01 \\
middle-4  & 0.42 & -0.09 & 0.29 \\
middle    & 0.39 & -0.09 & 0.00 \\
middle+4  & 0.38 & -0.12 & -0.02 \\
middle+8  & 0.38 & -0.12 & 0.07 \\
middle+12 & 0.38 & -0.13 & 0.06 \\
last      & 0.33 & -0.13 & 0.05 \\
\midrule
\rowcolor{light-gray}\multicolumn{4}{c}{Token aggregation} \\
Avg. response tokens  & 0.39 & -0.09 & 0.01 \\
Last response token   & 0.38 & -0.07 & -0.02 \\
Random response token & 0.21 & -0.12 & -0.01 \\
\midrule
Avg. all tokens       & 0.22 & -0.08 & 0.00 \\
Avg. context tokens   & 0.12 & -0.08 & -0.01 \\
Last context token    & 0.23 & 0.01 & 0.00 \\
\midrule
\rowcolor{light-gray}\multicolumn{4}{c}{Probe architecture} \\
SAPLMA (4 layers)     & 0.39 & -0.09 & 0.01 \\
Uhead (1 layer)       & 0.36 & -0.14 & 0.14 \\
Uhead (2 layers)      & 0.34 & -0.15 & -0.04 \\
\midrule
SAPLMA (2 layers)     & 0.39 & -0.15 & 0.10 \\
SAPLMA (3 layers)     & 0.38 & -0.11 & 0.01 \\
SAPLMA (5 layers)     & 0.42 & -0.09 & 0.07 \\
\bottomrule
\end{tabular}
\caption{PRR results (higher is better) for PubmedQA (long-form), averaged across Llama-3.1-8B and Gemma-2-9B. Each section varies one design dimension while holding others fixed. The layer section uses the SAPLMA architecture with average response token aggregation; the aggregation section uses the middle layer; the architecture section uses the middle layer with average response token aggregation.}
\label{tab:long_form_ablation}
\end{table}

\begin{table*}
\centering
\small
\begin{tabular}{l c c c c c}
\toprule
Method & ID & LOO & 1D-SameTask & DiffTask & 1D-DiffTask \\
\midrule
MSP                 & 0.57 & 0.57 & \textbf{0.57} & \textbf{0.57} & \textbf{0.57} \\
SAPLMA (mid)        & 0.63 & 0.46 & 0.26 & 0.35 & 0.32 \\
HBO (paper version) & \textbf{0.66} & \textbf{0.61} & \textbf{0.57} & \textbf{0.57} & \textbf{0.57} \\
HBO (simplified)    & 0.63 & 0.55 & \textbf{0.57} & \textbf{0.57} & \textbf{0.57} \\
\bottomrule
\end{tabular}
\caption{PRR results (higher is better) comparing the HBO configuration used in the paper against a simplified configuration ($UQ_{\text{usv}} = R$, $UQ_{\text{sv}} = 1-R$), averaged across all three short-form datasets (SciQ, TriviaQA, CoQA) for Llama 3.1-8B.}
\label{tab:hbo_simplified_all_short}
\end{table*}

\begin{table*}
\centering
\small
\begin{tabular}{l c c c c c}
\toprule
Method & ID & LOO & 1D-SameTask & DiffTask & 1D-DiffTask \\
\midrule
MSP                 & 0.58 & 0.58 & \textbf{0.58} & 0.58 & \textbf{0.58} \\
SAPLMA (mid)        & 0.63 & 0.57 & 0.27 & 0.35 & 0.35 \\
HBO (paper version) & \textbf{0.68} & \textbf{0.64} & \textbf{0.58} & 0.58 & \textbf{0.58} \\
HBO (simplified)    & 0.63 & 0.62 & \textbf{0.58} & \textbf{0.59} & \textbf{0.58} \\
\bottomrule
\end{tabular}
\caption{PRR results (higher is better) comparing the HBO configuration used in the paper against a simplified configuration, for SciQ only -- the dataset used as a validation set to select between the two configurations -- for Llama 3.1-8B.}
\label{tab:hbo_simplified_sciq}
\end{table*}

\begin{table*}
\centering
\small
\begin{tabular}{l c c c c c}
\toprule
Method & ID & LOO & 1D-SameTask & DiffTask & 1D-DiffTask \\
\midrule
MSP                 & 0.57 & 0.57 & \textbf{0.57} & \textbf{0.57} & \textbf{0.57} \\
SAPLMA (mid)        & 0.63 & 0.40 & 0.25 & 0.35 & 0.30 \\
HBO (paper version) & \textbf{0.66} & \textbf{0.59} & \textbf{0.57} & \textbf{0.57} & \textbf{0.57} \\
HBO (simplified)    & 0.63 & 0.51 & \textbf{0.57} & 0.56 & \textbf{0.57} \\
\bottomrule
\end{tabular}
\caption{PRR results (higher is better) comparing the HBO configuration used in the paper against a simplified configuration, averaged across the short-form datasets excluding SciQ (TriviaQA, CoQA) for Llama 3.1-8B.}
\label{tab:hbo_simplified_excl_sciq}
\end{table*}

\begin{table*}[t]
\centering
\small
\setlength{\tabcolsep}{4pt}
\begin{adjustbox}{width=\textwidth}
\begin{tabular*}{\textwidth}{@{\extracolsep{\fill}} l c c c c c c c c c}
\toprule
& \multicolumn{4}{c}{Degrees of OOD eval}
& \multicolumn{2}{c}{Breadth of eval}
& \multicolumn{2}{c}{Strong OOD baselines}
& \multicolumn{1}{c}{Claims} \\
\cmidrule(lr){2-5}\cmidrule(lr){6-7}\cmidrule(lr){8-9}\cmidrule(lr){10-10}
Method & None & LOO & 1D-SameTask & DiffTask & 3+ Eval & Nat. & Unsup. & Sup. & +ve claim \\
\midrule
\citet{vazhentsev-etal-2025-token} & & \cmark &  &  & \cmark & \cmark & \cmark & & \cmark \\ % DONE
\midrule
\citet{shelmanov-etal-2025-head} &   &  & \cmark &  &  \cmark & & \cmark &  & \cmark \\ % DONE
\midrule
\citet{obeso2025realtimedetectionhallucinatedentities} & &  & \cmark & \cmark & \cmark & \cmark & \cmark &  & \cmark  \\ % DONE
\midrule
%\midrule
\citet{chuang-etal-2024-lookback} &   &  &  \cmark & \cmark & \cmark & \cmark &  & \cmark &  \cmark \\ % TODO
\midrule
\citet{mckenzie2025detecting} &       &  &   & \cmark & \cmark & \cmark & & \cmark & \cmark \\
\midrule
\citet{he-etal-2024-llm} & & \cmark & & &  \cmark & &  & \cmark & \cmark \\ % Does not mention which layer was used for SAPLMA
\midrule
\citet{kapoor2024large} & & \cmark  & & \cmark & & \cmark & & & \cmark \\
\midrule
%\midrule
\citet{vazhentsev-etal-2025-unconditional} & & \cmark  &  & \cmark & \cmark & \cmark & \cmark &  &  \\ % DONE
\midrule
\citet{DBLP:conf/iclr/OrgadTGRSKB25} & &  & \cmark &  \cmark & \cmark & \cmark & \cmark &  &  \\ % DONE
\midrule
\citet{tong2025halunetmultigranularuncertaintymodeling} &  &  &  \cmark &  & \cmark  & \cmark  & \cmark &  &  \\% TODO
\midrule
\citet{kossen2024semanticentropyprobesrobust} &  & \cmark &  &  & \cmark & \cmark & \cmark &  \cmark &  \\ % DONE
%\midrule
\midrule
%\midrule
\citet{DBLP:conf/nips/BurgerHN24} & & \cmark  & & \cmark & \cmark & & & \cmark \\ 
%\midrule
\midrule
\citet{zhang2025reasoning} & & & \cmark & \cmark  & \cmark & \cmark  & & & \\ 
\midrule
\citet{huang-etal-2025-confidence} & \cmark \\
\midrule
\citet{su-etal-2024-unsupervised} & \cmark \\
\midrule
\citet{bhattacharjya-etal-2025-simba} & \cmark \\
\midrule
Ours (HBO) &  & \cmark & \cmark & \cmark & \cmark & \cmark & \cmark & \cmark & \cmark \\
\bottomrule
\end{tabular*}
\end{adjustbox}

\caption{Supervised UQ work which includes:  1) no OOD experimentation, 2) testing robustness in a leave-one-out setting (\textbf{\textit{LOO}}), 3) training on a single dataset for the same task (\textbf{\textit{SameTask}}), 4) training only on different tasks (\textbf{\textit{DiffTask}}), 5) containing OOD experimentation beyond synthetically generated data (\textbf{\textit{Nat.}}), 6) testing on 3 or more OOD datasets (\textbf{\textit{3+ eval}}), 7) using a logit-based baseline such as MSP or entropy (\textbf{\textit{Unsup.}}), and 8) including a strong supervised baseline where linear or MLP probes are trained only on the middle hidden state of an LLM (\textbf{\textit{Sup.}}). \textbf{+ve claim} refers to methods that on average outperform each of the baselines they were compared to when tested OOD.} 
\label{tab:survey_ood_experimentation_all_papers}
\end{table*}

% UNCOMMENT FOR ARR
%\section{ARR Check-list Information} 

%\paragraph{AI usage.}
%Claude and ChatGPT were used to provide feedback on the writing in this draft, and for help refactoring the project code.

%\paragraph{Hyper-parameters and prompts.} We use the same prompts and base model hyper-parameters as those used by \citet{vazhentsev-etal-2025-token}, allowing for direct comparability of our leave-one-out evaluation. For hyper-parameter tuning for the probe methods, we use the hyper-parameters chosen in prior work (for example, we show results using both sets of hyper-parameters proposed by \citet{shelmanov-etal-2025-head}). Our HBO method does not require additional hyper-parameter tuning.

\end{document}